\documentclass{WileyMSP-template}
\usepackage{amsmath, amsfonts}
\usepackage{algorithmic}
\usepackage{algorithm}
\usepackage{array}
\usepackage[caption=false, font=normalsize, labelfont=sf, textfont=sf]{subfig}
\usepackage{textcomp}
\usepackage{stfloats}
\usepackage{url}
\usepackage{verbatim}
\usepackage{graphicx}
\usepackage{cite}
\usepackage{color}
\usepackage{titlecaps}
\usepackage{mathcomp}
\usepackage{bm}

\newcommand{\secref}[1]{Section \ref{#1}}
\newcommand{\subsecref}[1]{Section \ref{#1}}
\newcommand{\tabref}[1]{{Table \ref{#1}}}
\newcommand{\figref}[1]{{Fig. \ref{#1}}}
\newcommand{\equref}[1]{{(\ref{#1})}} 

\newcommand{\circletext}[1]{\raisebox{.5pt}{\textcircled{\raisebox{-.5pt} {{\footnotesize #1}}}}}

\begin{document}

\pagestyle{fancy}
\rhead{\includegraphics[width=2.5cm]{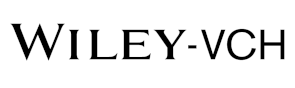}}

\title{REWW-ARM -- Remote Wire-Driven Mobile Robot:\\ Design, Control, and Experimental Validation}
\maketitle

\author{Takahiro Hattori}
\author{Kento Kawaharazuka}
\author{Temma Suzuki}
\author{Keita Yoneda}
\author{Kei Okada}


\begin{affiliations}
Takahiro Hattori, Kento Kawaharazuka, Temma Suzuki, Keita Yoneda, Kei Okada \\
Department of Mechano-Informatics, Graduate School of Information Science and Technology, The University of Tokyo, 7-3-1 Hongo, Bunkyo-ku, Tokyo, 113-8656, Japan \\
Email Address :  [t-hattori, kawaharazuka, t-suzuki, yoneda, k-okada]@jsk.imi.i.u-tokyo.ac.jp \\

Kento Kawaharazuka \\
AI Center, Graduate School of Information Science and Technology, The University of Tokyo, Japan \\

\end{affiliations}



\begin{abstract}
Electronic devices are essential for robots but limit their usable environments. To overcome this, methods excluding electronics from the operating environment while retaining advanced electronic control and actuation have been explored. These include the remote hydraulic drive of electronics-free mobile robots, which offer high reachability, and long wire-driven robot arms with motors consolidated at the base, which offer high environmental resistance. To combine the advantages of both, this study proposes a new system, "Remote Wire Drive." As a proof-of-concept, we designed and developed the Remote Wire-Driven robot "REWW-ARM", which consists of the following components: 1) a novel power transmission mechanism, the "Remote Wire Transmission Mechanism" (RWTM), the key technology of the Remote Wire Drive; 2) an electronics-free distal mobile robot driven by it; and 3) a motor-unit that generates power and provides electronic closed-loop control based on state estimation via the RWTM. In this study, we evaluated the mechanical and control performance of REWW-ARM through several experiments, demonstrating its capability for locomotion, posture control, and object manipulation both on land and underwater. This suggests the potential for applying the Remote Wire-Driven system to various types of robots, thereby expanding their operational range.
\end{abstract}


\section{\titlecap{Introduction}}\label{sec:introduction}
One of the most important roles required of robots is to operate in dangerous and harsh environments in place of humans. For example, the DARPA Robotics Challenge\cite{DARPA_robotics_challenge} and the ImPACT: Tough Robotics Challenge \cite{impact} were themed on the introduction of robots into harsh disaster environments that are inaccessible to humans. Major humanoid and quadruped robots such as Atlas\cite{atlas} and ANYmal\cite{anymal} also position the substitution of human work in dangerous and hard-to-enter places as one of their key missions.
However, electronic components such as motors and sensors, which are indispensable for controlling robots, are vulnerable to such environments. Factors that prevent human entry, such as moisture, radiation, high and low temperatures, dust, flammable gases, and high voltages, also damage precision electronic components \cite{hazard_protection, electronics_harsh, vimala2009corrosion, zhao2023researchCorrosion, almubarak2017heatAndElectronics, han2016protectionPassiveHeat, daneshvar2021multilayerRadiation, martin2009protectingMoisture}.

\begin{figure}
\centering
\includegraphics[width=0.9\linewidth]{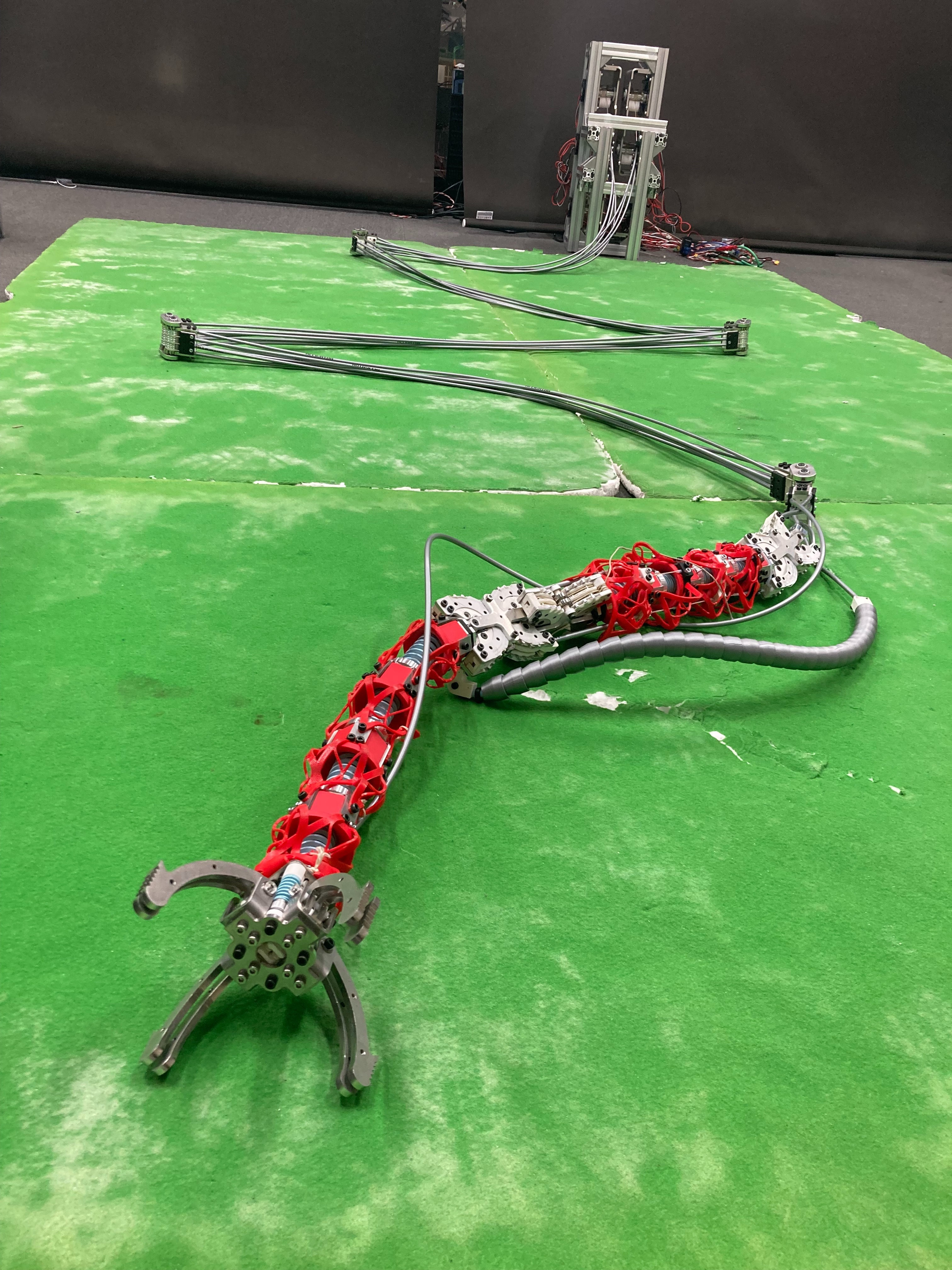}
\caption{Overview of REWW-ARM, the Remote Wire-Driven robot developed in this study. Power is physically transmitted from the motor-unit in the background to the snake-like robot in the foreground through the slender Remote Wire Transmission Mechanism.}
\vspace{-2.0ex}
\label{fig:fig1}

\end{figure}

One way to address this dilemma is to use electronics-free autonomous mobile robots. For example, the robot developed by Drotman et al. \cite{electronics_free_air_powered} is a three-legged robot that drives pneumatic artificial muscles with a pneumatic circuit, capable of walking and changing direction upon contact detection. The modular robot developed by He et al. \cite{modular_electfree} enables decision and action without electronic devices by incorporating materials that deform in response to chemicals, light, and heat into an origami structure. Furthermore, the robot developed by Stephen et al. \cite{extreme_electfree_soft} is a six-legged pneumatic robot that is autonomously controlled by transistors that switch fluids instead of electricity, giving it high explosion-proof properties.
However, these non-electronic information processing mechanisms have significantly lower computational power per unit volume and mass compared to their electronic counterparts, making advanced decision and action difficult. Another issue is that non-electric actuators have significantly inferior controllability compared to electric actuators such as motors.

Therefore, another approach is to separate the electronic devices from the robot's moving parts via a transmission medium. By transmitting power from electronic devices consolidated at the base to the moving parts at the tip via a physical medium, it is possible to achieve both the exclusion of electronic components from the operating environment and electronic control and actuation.

One such method using a medium is to remotely drive an electronics-free mobile robot. For example, the nuclear power plant work robot HUMALT developed by Nakamura et al. \cite{HUMALT_RSJ} remotely drives an electronics-free hydraulically driven mobile robot through a several-tens-of-meters-long tube from a pump and control unit. However, water has low temperature resistance, boiling at high temperatures and freezing at low temperatures. Hydraulic systems also involve mechanical constraints and complexities such as pumps, valves, and watertight structures.

Another method of separating electronic devices and robots using a medium is a long wire-driven robot arm. For example, the 10-meter-long wire-driven robot arm for nuclear power plant work, Super Dragon, developed by Endo et al. \cite{super_dragon}, achieves a lighter arm and improved environmental resistance by consolidating the actuators at the base. Similarly, the 2.3-meter-long continuum robot arm developed by Qin et al. \cite{fusion_snake_arm} also has its actuators consolidated at the base, giving it high radiation resistance for expected use in fusion reactors. Other recent studies have also explored wire-driven systems for various applications, such as continuum manipulators with online error compensation \cite{0339} and multi-modal soft robots \cite{0204}. In general, wires have the advantages over water of not leaking, thus not requiring sealing, and having a wider usable temperature range. For example, the wire jamming mechanism gripper developed by Tadakuma et al. \cite{wirejamming} has high fire resistance. The chemical fiber Vectran\cite{vectran, vectran_about}, used in various wire-driven robots\cite{SAQIEL, Cubix}, is also capable of operating in a wide temperature range from -70\textdegree C to 400\textdegree C. On the other hand, these robot arms have limited reachability and obstacle avoidance capabilities compared to mobile robots.

Therefore, we propose a novel system, the Remote Wire Drive, which can make advanced decision and action unlike electronics-free autonomous mobile robots, and combines the high reachability of remote driving with the high environmental resistance of wire-driven system. A Remote Wire Drive is a system in which electronic devices are separated from a mobile robot, consolidated externally, and power is transmitted through wires. In this study, we developed "REWW-ARM," shown in \figref{fig:fig1}, as such a Remote Wire-Driven robot. REWW-ARM consists of the following three elements: 1) a motor-unit that consolidates motors and electronic devices; 2) a special mechanism, the "Remote Wire Transmission Mechanism (RWTM)," that enables power transmission from the motor-unit through wires; and 3) an arm-like snake mobile robot at the distal end that is driven by it and moves freely. The distal mobile robot receives power transmission from the RWTM and also transmits its state to the motor-unit via the RWTM. This enables advanced movements based on closed-loop control, despite containing no electronic devices at all. In this study, we develop the hardware and software for REWW-ARM and verify the feasibility of the Remote Wire-Driven system by evaluating its capabilities through several experiments.

\section{\titlecap{Mechanical Design}}\label{sec:design}
\begin{figure}
    \centering
    \includegraphics[width=1.0\linewidth]{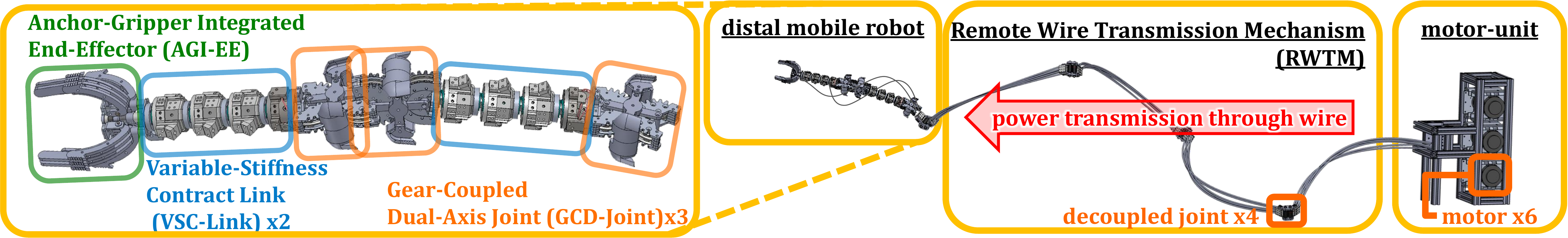}
        \caption{Overview of design. REWW-ARM consists of three elements: a motor-unit that generates power, a Remote Wire Transmission Mechanism (RWTM) that transmits power, and a distal mobile robot that is driven by the transmitted power. The distal mobile robot is further composed of three Gear-Coupled Dual-Axis Joints (GCD-Joints), two Variable-Stiffness Contract Links (VSC-Links), and one Anchor-Gripper Integrated End-Effector (AGI-EE), which cooperate to enable locomotion and object manipulation.}
    \vspace{-2.0ex}
    \label{fig:concept}
\end{figure}
This section describes the mechanical design of REWW-ARM. An overview is shown in \figref{fig:concept}. REWW-ARM is broadly composed of the following three elements:
\begin{itemize}
    \item \textbf{motor-unit}: This unit houses the motors, computer, and other electronic devices, and is responsible for generating power and controlling REWW-ARM.
    \item \textbf{Remote Wire Transmission Mechanism (RWTM)}: This mechanism transmits power from the motor-unit to the distal mobile robot via wires. It has a special structure that allows it to transmit wire tension without hindering the free movement of the distal mobile robot.
    \item \textbf{distal mobile robot}: An electronics-free, arm-like snake robot powered by the motor-unit through the RWTM. It consists of three Gear-Coupled Dual-Axis Joints (GCD-Joints), two Variable-Stiffness Contract Links (VSC-Links), and one Anchor-Gripper Integrated End-Effector (AGI-EE), enabling both locomotion and object manipulation.
\end{itemize}
This section will describe these elements in order.
The specifications are shown in \tabref{tab:specifications}.

\begin{table}[htbp]
    \centering
    \caption{Specifications of REWW-ARM}
    \begin{tabular}{|p{4.8cm}|p{3cm}|} 
        \hline
        \textbf{item} & \textbf{value} \\ 
        \hline
        the number of decoupled joints & 4 \\ \hline
        decoupled joint angle range & $\pm 135$ \textdegree\\ \hline
        total length of RWTM & 4 m \\ \hline
        number of joints in distal mobile robot & 3\\ \hline
        GCD-Joint angle range & $\pm 130$ \textdegree\\ \hline
        VSC-Link length range & 0.185$\sim$ 0.302 m \\ \hline
        number of wires & 6 \\ \hline
        number of motors & 6 \\ \hline
        motor model & Steadywin GIM8108-8 \\ \hline
        winding winch diameter & 0.019 m\\ \hline
        maximum continuous wire tension & 500 N\\ \hline
        wire elongation per unit tension & 0.000185 m/N\\ \hline
        maximum continuous joint torque (independent per joint) & 11.6 N$\cdot$ m\\ \hline
        maximum continuous joint torque (not independent per joint) & 23.2 N$\cdot$ m\\ \hline
    \end{tabular}
    \vspace{-2.0ex}
    \label{tab:specifications}
\end{table}

\subsection{\titlecap{motor-unit}}
\begin{figure}
    \centering
    \includegraphics[width=0.5\linewidth]{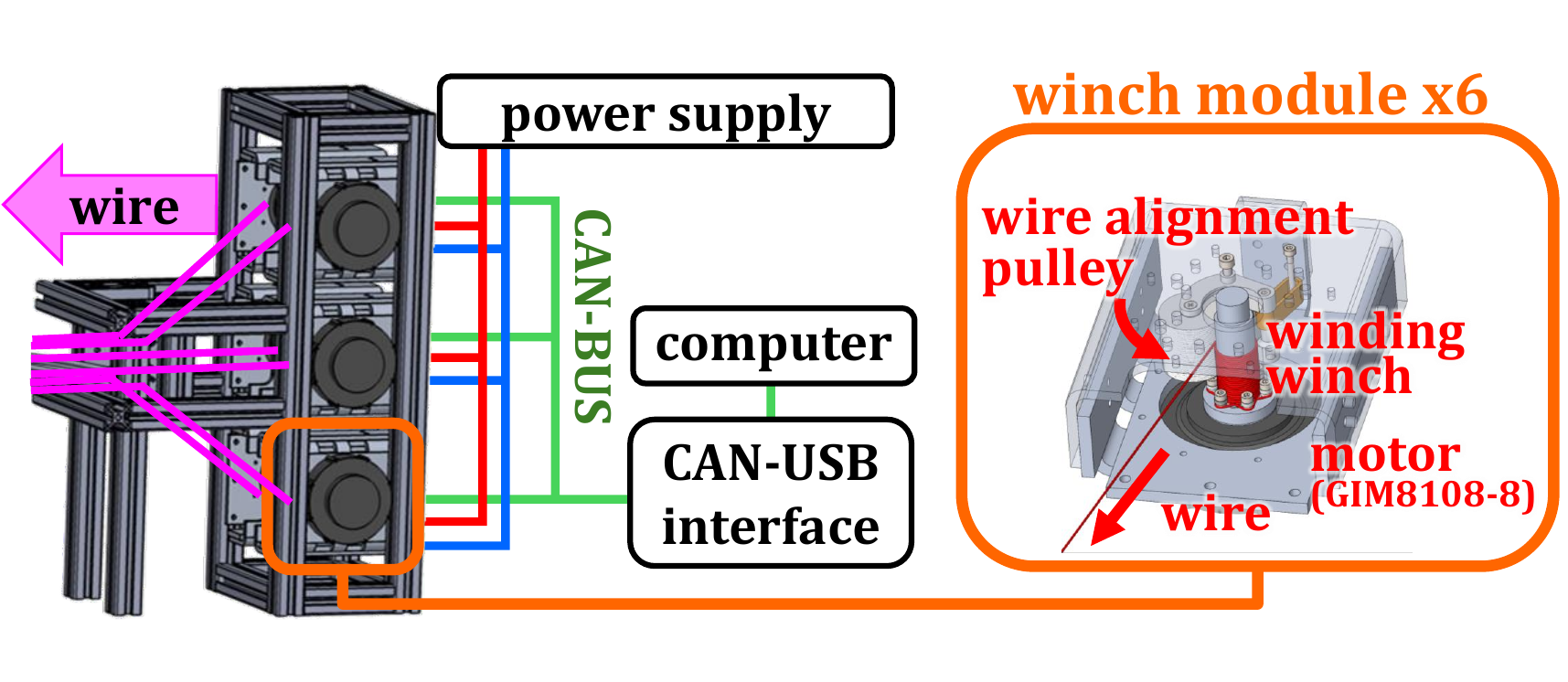}
        \caption{The motor-unit consists of six winch modules controlled by CAN communication, controlling six wires.}
    \vspace{-2.0ex}
    \label{fig:motor_unit}
\end{figure}
The motor-unit, as shown in \figref{fig:motor_unit}, consists of six winch modules that pull the wires, and is the only part of REWW-ARM that contains electronic systems. Each winch module winds a wire using a winding pulley on the motor shaft. A wire alignment pulley is pressed by a spring to ensure the wire is wound neatly. The motors are controlled by a computer via CAN communication and are supplied with power from a power supply unit.

\subsection{\titlecap{Remote Wire Transmission Mechanism}}
\label{subsec:RWTM_design}

\begin{figure}
    \centering
    \includegraphics[width=0.5\linewidth]{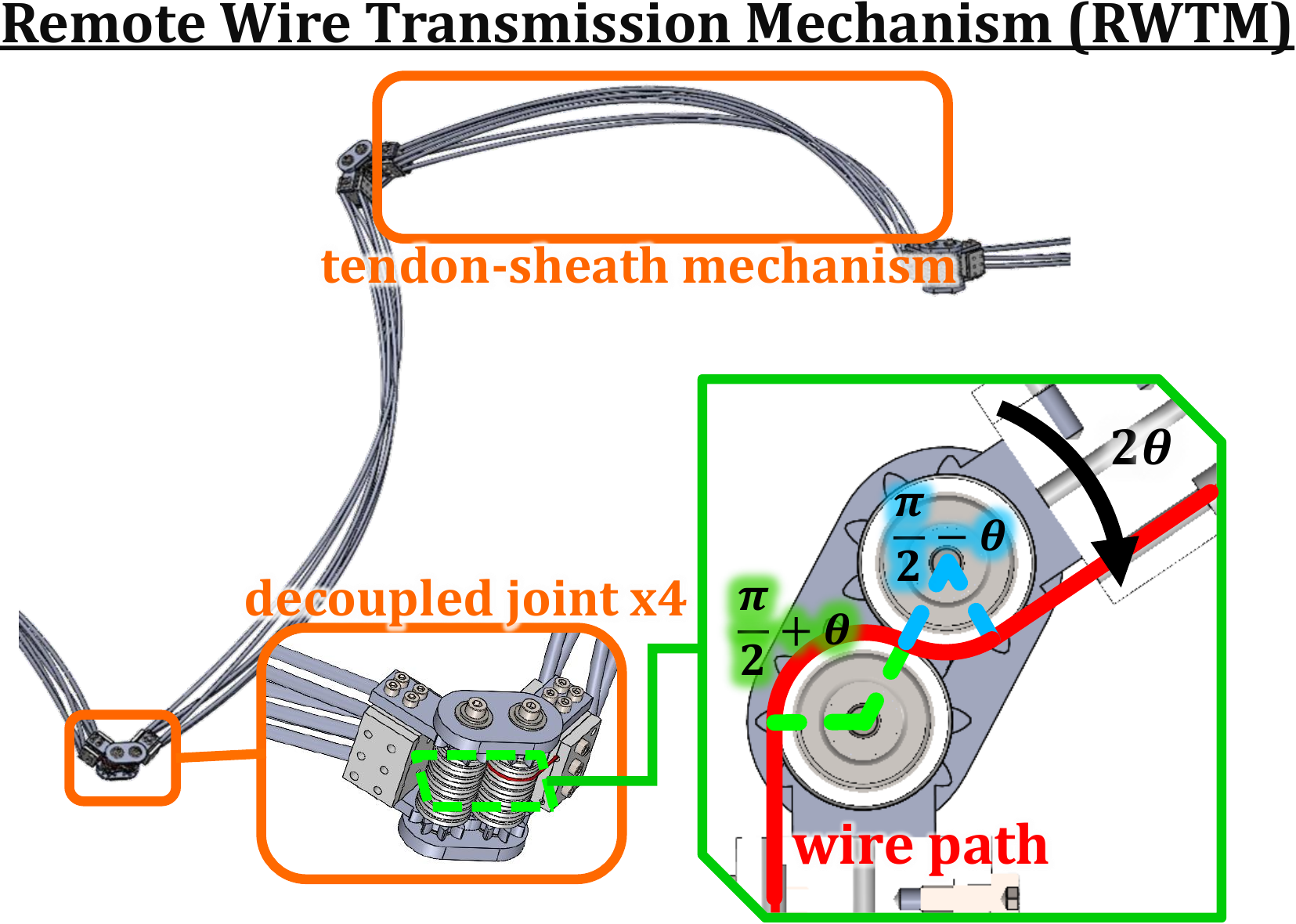}
        \caption{The RWTM has a structure in which tendon-sheath mechanisms and decoupled joints are alternately connected in series. The high flexibility of the tendon-sheath mechanism and the high and stable transmission efficiency of the decoupled joint complement each other, enabling stable and highly efficient transmission without hindering the movement of the distal mobile robot.}
    \vspace{-2.0ex}
    \label{fig:RWTM}
\end{figure}

This section describes the Remote Wire Transmission Mechanism (RWTM). This is a mechanism that transmits power via wires from the motor-unit to the distal mobile robot without generating forces that would hinder the robot's movement.

\begin{figure}%
    \centering
    \includegraphics[width=0.5\linewidth]{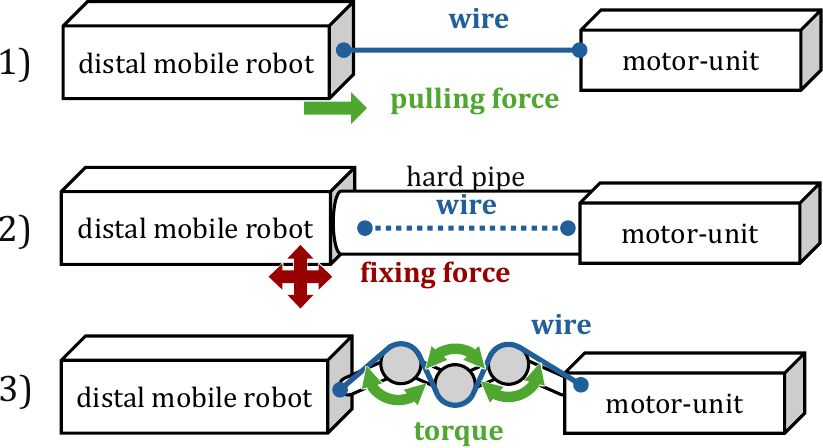}
        \caption{Examples of wire mechanisms that do not work as an RWTM. External forces are applied to the distal mobile robot due to wire tension or lack of flexibility, hindering its movement.}
    \vspace{-2.0ex}
    \label{fig:RWTM_nonsuitable}
\end{figure}

To realize the RWTM, it is not enough to simply connect the wires from the motor-unit to the distal mobile robot; several design requirements must be met. To clarify this, unsuitable examples for an RWTM are shown in \figref{fig:RWTM_nonsuitable}, and explained below.
\begin{enumerate}
    \item For example, if you simply connect a wire from the motor-unit to the distal mobile robot, a pulling force is applied to the distal mobile robot, hindering its movement.
    \item If the wire is passed through a rigid tube to counteract the wire tension, the distal mobile robot is fixed in place and cannot move.
    \item When using a serial link with pulleys placed on the axis, the wire tension generates joint torque \cite{Coupled_tendon_driven}, which also produces an external force that hinders the autonomous movement of the distal mobile robot.
\end{enumerate}

From the above, it is clear that the RWTM must satisfy the following conditions:
\begin{itemize}
    \item It must not deform due to wire tension or displacement, and must not generate external forces on the distal mobile robot;
    \item Regardless of wire tension or displacement, it must deform flexibly in response to external forces caused by the movement of the distal mobile robot.
\end{itemize}

Two mechanisms that satisfy these conditions are the tendon-sheath mechanism (TSM) and the decoupled joint. In this study, we constructed the RWTM by connecting these two alternately in series, as shown in \figref{fig:RWTM}. Below, we will explain these two mechanisms and then state the reason for combining them.

The TSM \cite{tendon_sheath_robot} is a mechanism used in bicycle brake cables, surgical instruments \cite{tendon_sheath_actuated_manipulator}, and wearable robots \cite{RAS_hand, RAS_for_Wearable}. A wire passes through a flexible metal tube with a spring-like structure, the sheath, whose inner wall is coated with a low-friction material.

A decoupled joint is a joint mechanism with the property that joint torque and wire tension, as well as joint displacement and wire displacement, are independent.
Normally, a joint mechanism that includes a wire path has a coupled relationship between joint displacement and wire displacement \cite{Coupled_tendon_driven}. However, a decoupled joint, as shown in \figref{fig:RWTM}, is a double joint in which two axes are synchronized in opposite directions by a gear. The two axes cancel out the coupling relationship, so the displacement and force of the joint and wire are independent of each other. This special joint has been used in robot arms such as D3-ARM \cite{D3_arm} and LIMS \cite{LIMS2}. However, its purpose was to simplify the calculation of the coupling relationship, not to be used as a power transmission mechanism as in this study.

Comparing the properties of these two mechanisms, we find the following.
\begin{itemize}
    \item \textbf{Friction coefficient:}
    In TSM, the inner wall of the sheath and the wire are in sliding contact, so the friction coefficient is high, ranging from 0.04 to 0.2 even when using PTFE, a material known for its low friction, on the inner wall \cite{bowden_friction}. On the other hand, in a decoupled joint, the wire only passes over pulleys with bearings, so the friction coefficient is about 0.001 \cite{bearing_friction}.
    \item \textbf{Relationship between angle and transmission efficiency:}
    In TSM, the transmission efficiency $\eta$ decreases exponentially with respect to the cumulative bending angle as shown in \equref{eq:capstan_friction} \cite{bowden_friction}. Here, $\mu$ is the friction coefficient and $\alpha$ is the cumulative bending angle.
    \begin{equation}
        \eta = \exp (-\mu \alpha)
        \label{eq:capstan_friction}
    \end{equation}
    On the other hand, in a decoupled joint, the total wrap angle is constant regardless of the angle, and the transmission efficiency is also constant. The wire diameter used this time is 1 mm, and the pulley diameter is 20 mm. Therefore, based on the previous study by Suzuki et al. \cite{wiretester}, the transmission efficiency $\eta_\textrm{dj}$ per decoupled joint is estimated to be about 98-99\%.

    Comparing the transmission efficiencies of the decoupled joint and TSM based on this information, we get \figref{fig:efficiency_compare}, which shows that the decoupled joints have higher transmission efficiency in most of the range. The proportion of this range can be calculated as follows.
    
    The decoupled joint used in this study has a range of motion of $\pm {3 \pi}/{4}$ rad. Therefore, any $n$ decoupled joints connected in series can handle a cumulative bending angle of ${3\pi n}/{4}$ rad. The transmission efficiency $\eta$ of $n$ series-connected decoupled joints in the angle range $0 \leq \alpha \leq 3\pi n / 4$ is as shown in \equref{eq:eff_comp}.
    \begin{equation}
        \eta=\eta_\textrm{dj} ^ n
        \label{eq:eff_comp}
    \end{equation}
    Comparing \equref{eq:capstan_friction} and \equref{eq:eff_comp}, the interval where the efficiency of the decoupled joints is higher, $\eta_\textrm{dj} ^ n > \exp({- \mu \alpha})$, is equivalent to $({n}/{\mu}) \ln({1}/{\eta_\textrm{dj}}) < \alpha $, and the proportion within the angle range $0 \leq \alpha \leq {3 \pi n}/{4}$ is ${4\ln({1}/{\eta_\textrm{dj}})}/({3\pi\mu}) $. Calculating this for $0.98 \leq \eta_\textrm{dj} \leq 0.99$ and $0.04 \leq \mu \leq 0.2$, we get 0.79 to 0.98. In other words, regardless of the number of joints $n$, the transmission efficiency of the decoupled joints is superior in the angle range of 78\% to 98\%.
    \begin{figure}%
        \centering
        \includegraphics[width=0.5\linewidth]{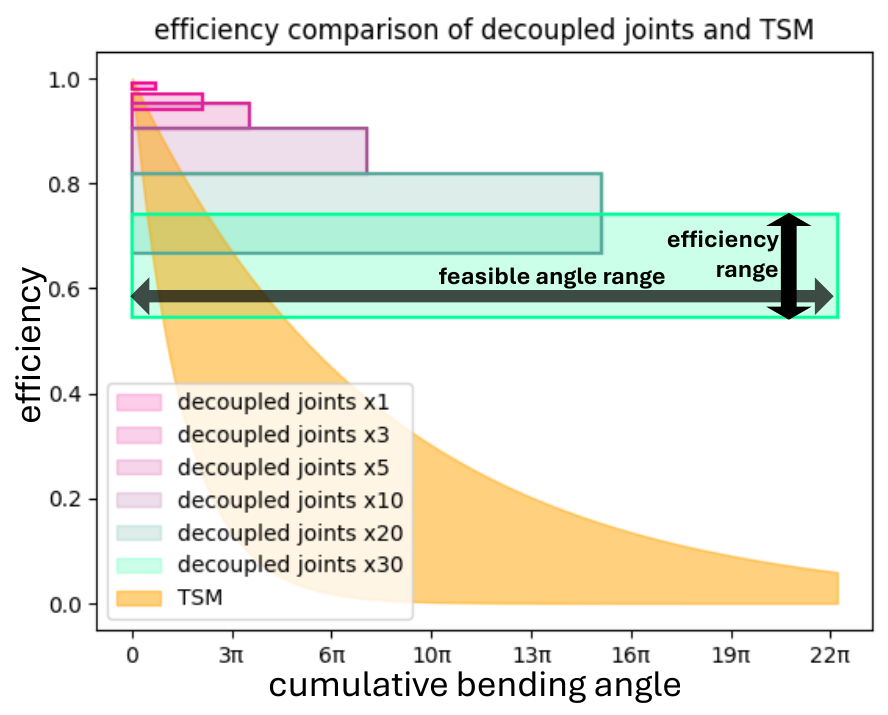}
                \caption{Comparison of transmission efficiency between TSM and decoupled joints. The maximum angle range and theoretical transmission efficiency range for various numbers of joints are shown as rectangles, superimposed on the theoretical transmission efficiency of TSM. For any number of joints, it shows higher and more stable transmission efficiency than TSM over most of the corresponding angle range.}
        \vspace{-2.0ex}
        \label{fig:efficiency_compare}
    \end{figure}
    \item \textbf{Degrees of freedom:}
    A decoupled joint can only bend in a specific direction on a specific axis. Also, due to its structure as shown in \figref{fig:RWTM}, it is not suitable for realizing a compact torsional degree of freedom. On the other hand, a TSM of sufficient length can perform deformations such as torsion and bending at any point. In particular, the torsional degree of freedom has almost no effect on transmission efficiency.
    \item \textbf{Elasticity:}
    In TSM, deformation and wire displacement are independent, but the sheath has elasticity, so it produces a slight restoring force against deformation. On the other hand, a decoupled joint does not produce a restoring force to a specific angle, so it can be said to have high compliance to external forces.
\end{itemize}

As described above, TSM and decoupled joints each have their own unique advantages and disadvantages. We developed the RWTM by connecting them alternately in series as shown in \figref{fig:RWTM} in order to take advantage of the benefits of both TSM and decoupled joints and to compensate for their respective shortcomings.
When a bending external force is applied to this RWTM, the decoupled joint preferentially bears the bending angle instead of the elastic TSM. And for bending external forces of arrangements and numbers that cannot be handled by the decoupled joints, or for torsional external forces, the TSM deforms. By deforming in these two stages, it is possible to achieve both the advantages of the decoupled joint, which are high transmission efficiency and high compliance, and the advantages of the TSM, which are high deformation freedom and compatibility with torsional degrees of freedom.

\subsubsection{\titlecap{Wire Material: Vectran Fiber}}

The choice of wire material is critical for the performance and durability of a wire-driven system, especially for applications in harsh environments. In this study, we use Vectran, a high-performance liquid-crystal polymer fiber. Vectran offers several advantages that make it highly suitable for Remote Wire Drive applications. It exhibits excellent flex-fatigue resistance; for instance, a 2mm diameter Vectran rope can withstand over 41,909 bending cycles, which is 1.6 to 20 times more than other high-strength fibers like PBO and aramid \cite{vectran_flex-fatigue}. Furthermore, Vectran boasts superior specific strength and specific modulus compared to metals, as well as high impact, abrasion, chemical, and radiation resistance \cite{vectran_official}. These properties ensure the longevity and reliability of the transmission system, which is crucial for robots operating in extreme environments.

\subsection{\titlecap{distal mobile robot}}
\label{subsec:robotpartdesign}
\begin{figure}
    \centering
    \includegraphics[width=0.7\linewidth]{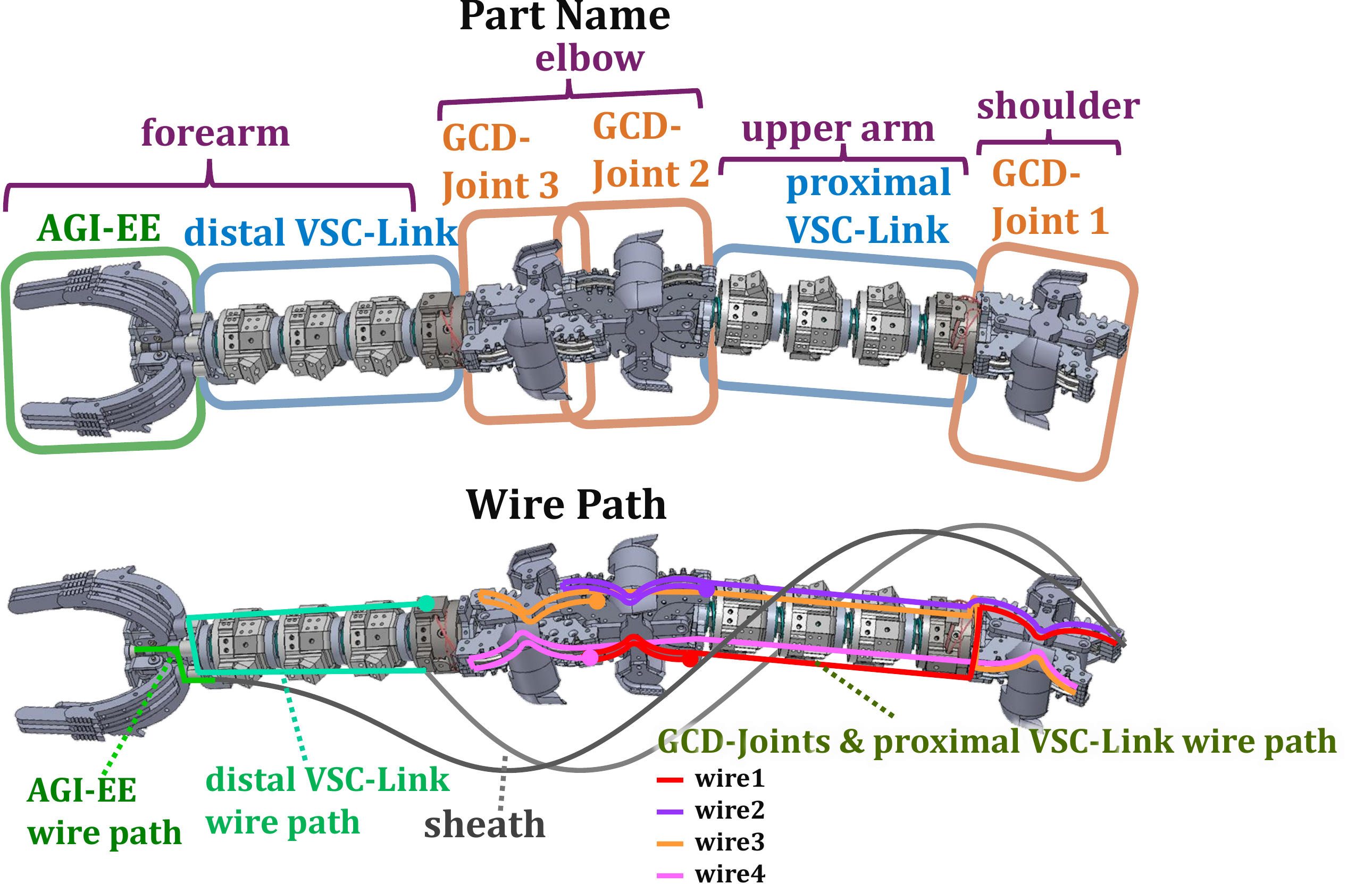}
        \caption{The parts of the distal mobile robot are named shoulder, upper arm, elbow, and forearm in order from the base. There are a total of six wires, four of which are used for the coupled wire drive of the shoulder, upper arm, and elbow. The remaining two wires are used to drive the distal VSC-Link and AGI-EE, respectively.}
    \vspace{-2.0ex}
    \label{fig:mobile_robot_part}
\end{figure}

This section describes the distal mobile robot. The distal mobile robot is an arm-like snake robot as shown in \figref{fig:concept} and \figref{fig:mobile_robot_part}, and consists of three Gear-Coupled Dual-Axis Joints (GCD-Joints), two Variable-Stiffness Contract Links (VSC-Links), and one Anchor-Gripper Integrated End-Effector (AGI-EE). To verify the broad effectiveness of the Remote Wire-Driven system, we made it capable of not only autonomous locomotion but also object manipulation. The names of each part are shown in \figref{fig:mobile_robot_part}.

The design requirements are as follows.
\begin{itemize}
    \item \textbf{The tendon Jacobian should be almost constant regardless of the angle}: 
    In a coupled wire-driven robot, the wire length $\bm{l}$ and the joint angle $\bm{q}$ satisfy the relationship shown in \equref{eq:tendon_len_eq_der} using a matrix called the tendon Jacobian $\bm{G}$ \cite{Coupled_tendon_driven}. Here, $\bm{G}\{i, j\}$ is the moment arm of wire i around joint j.
    \begin{equation}
        \dot{\bm{l}} = \bm{G(q)} \dot{\bm{q}}
        \label{eq:tendon_len_eq_der}
    \end{equation}
    Since the distal mobile robot does not have angle sensors on its joints, it estimates the joint angles from the wire lengths. It is desirable for this estimation to be a linear problem so that it can be calculated stably. Therefore, it is desirable to have an arc-shaped joint wire path where the tendon Jacobian is constant regardless of $\bm{q}$.
    \item \textbf{It should be able to move autonomously with simple control}: 
    There are various methods of locomotion for snake robots \cite{snake_robot_locomotion, snake_robot_tutorial}. However, it is difficult to create a mechanism that rotates indefinitely like a wheel in a wire-driven robot. On the other hand, locomotion methods that require a large number of joints, such as meandering, are not suitable for this robot. Therefore, we adopted peristalsis, which can be achieved with discrete control, as the main means of locomotion.
\end{itemize}
Based on these design requirements, we manufactured the GCD-Joint, VSC-Link, and AGI-EE.

\subsubsection{\titlecap{GCD-Joint}}
\begin{figure}
    \centering
    \includegraphics[width=0.5\linewidth]{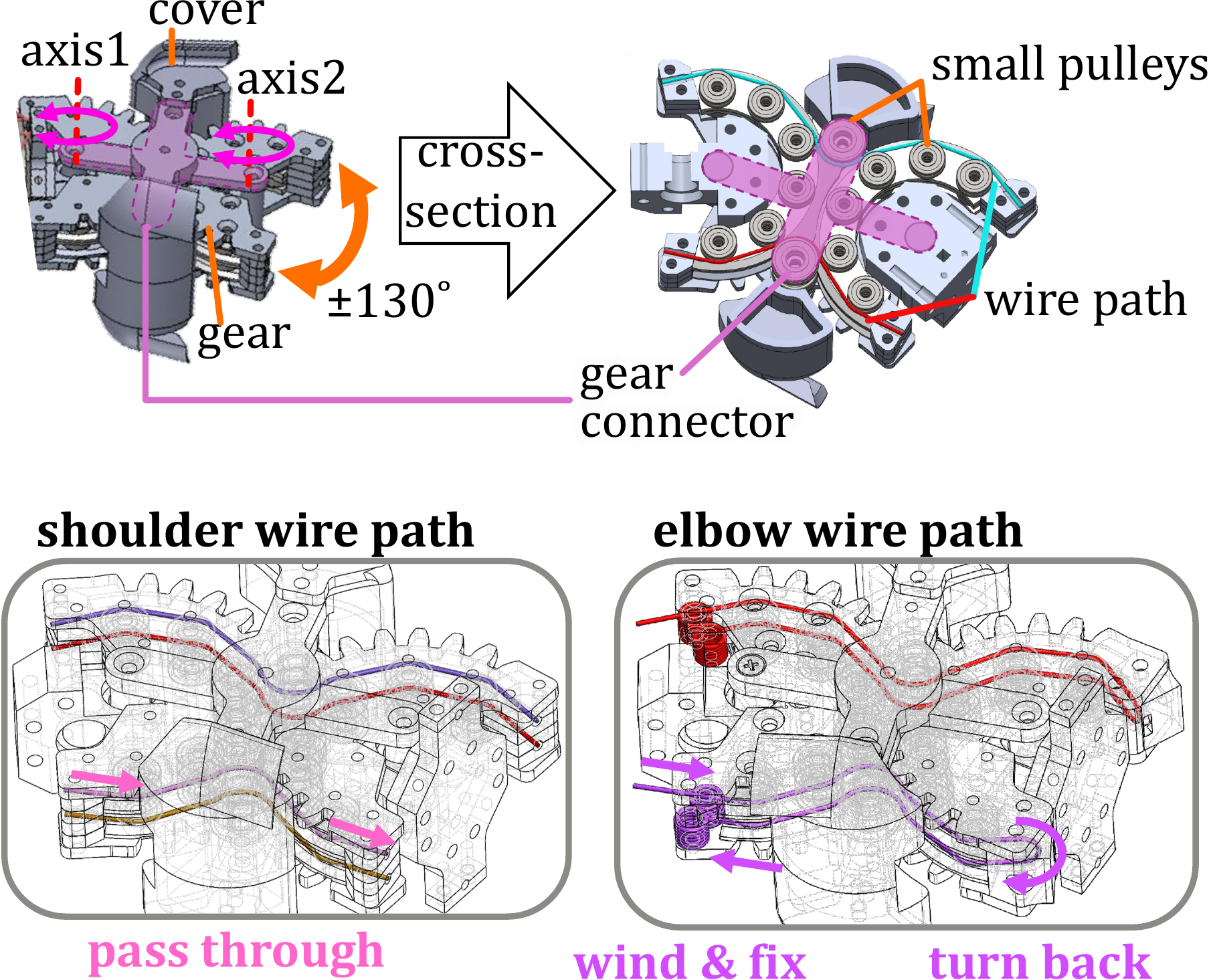}
        \caption{The GCD-Joint is a dual gear joint with a high range of motion. By arranging small-diameter pulleys on the circumference, it achieves both compactness and stability of the moment arm. Two types of wire paths can be realized depending on whether there is a turn-back, and turning back results in a 2x reduction. This allows for adjustment to increase the symmetry of the torque exerted by each joint.}
    \vspace{-2.0ex}
    \label{fig:dual_joint}
\end{figure}

The joint is a dual-joint structure with a range of motion of about $\pm 130$ degrees, synchronized by gears.
Small pulleys are arranged on the concentric circumference of a gear with a pitch radius of 30 mm, as shown in \figref{fig:dual_joint}. This results in a compact semicircular shape, yet the change in the moment arm is only about 4\%, which can be said to meet the design requirements.
Also, the tendon Jacobian is as shown in \equref{eq:tendon_Jacobian_val}.
\begin{equation}
    \bm{G}=\left(
    \begin{array}{cccc}
        0.028 & 0.056 & 0\\
        0.028 & -0.056 & 0 \\
        -0.028 & 0 & 0.056\\
        -0.028 & 0 & -0.056 
    \end{array}
    \right)
    \label{eq:tendon_Jacobian_val}
\end{equation}

\begin{figure}
    \centering
    \includegraphics[width=0.7\linewidth]{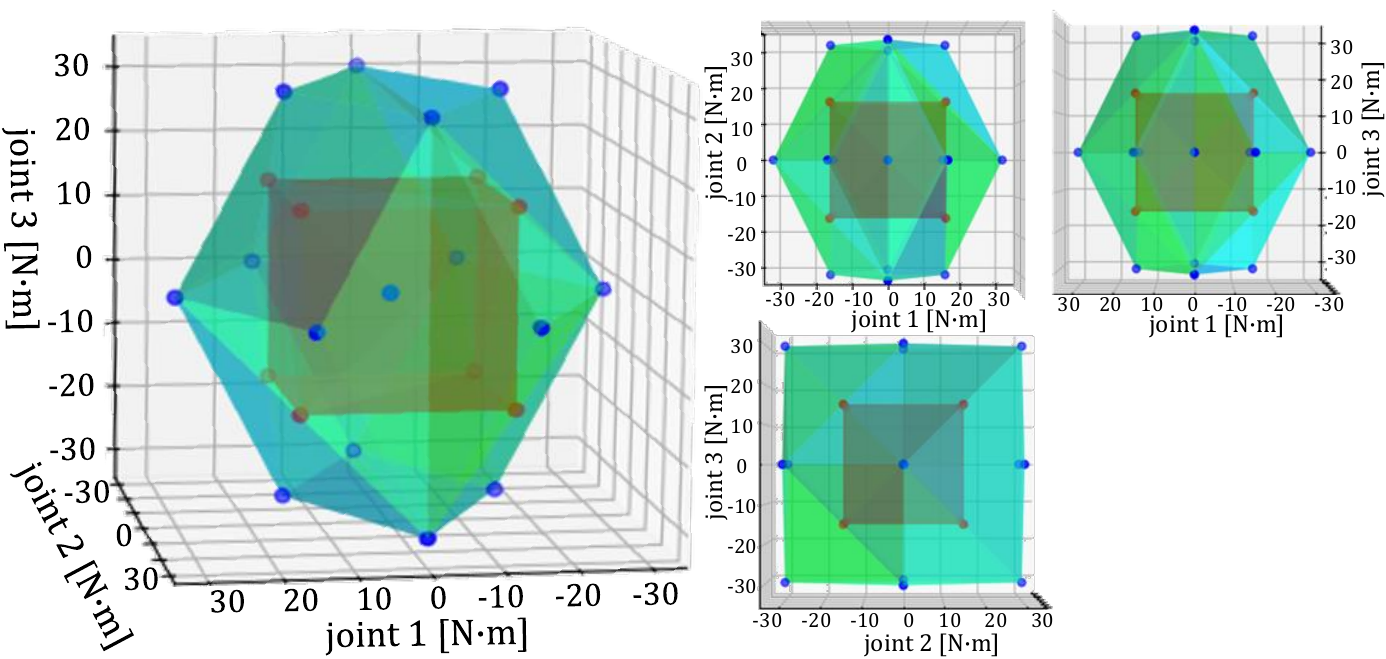}
        \caption{Exerted torque range. The polyhedron range in blue-green in the figure is the torque range, and the red cube inside it is the range of torque that can be exerted independently by the joints.}
    \vspace{-2.0ex}
    \label{fig:torque_range}
\end{figure}

Here, the torque range can be calculated from the relationship between the joint torque $\bm{\tau}$ and the wire tension $\bm{F}$ in \equref{eq:tendon_torque_eq}. This equation is a well-established relationship in wire-driven robotics, derived from the principle of virtual work and \equref{eq:tendon_len_eq} under the assumption that friction at the joints is negligible. It represents the geometric transformation from wire tensions to joint torques. The resulting torque space shape is highly symmetrical as shown in \figref{fig:torque_range}. Its maximum value is as shown in \tabref{tab:specifications}.
\begin{equation}
    \bm{\tau} = - \bm{G}^T \bm{F}
    \label{eq:tendon_torque_eq}
\end{equation}
\subsubsection{\titlecap{VSC-Link}}
The design requirements for the VSC-Link are as follows.
\begin{itemize}
    \item It must be able to expand and contract to satisfy the condition of being able to perform peristaltic motion.
    \item It must be able to become flexible when extended. This allows the AGI-EE to be robustly grounded against joint angle errors, enabling stable peristaltic motion. The role of the AGI-EE in peristaltic motion will be explained in detail in \secref{subsubsec:AGI-EE} and \secref{subsec:proceed_control}.
    \item It must also be able to become rigid so as not to absorb the intended movement of the joints.
\end{itemize}

\begin{figure}
    \centering
    \includegraphics[width=0.5\linewidth]{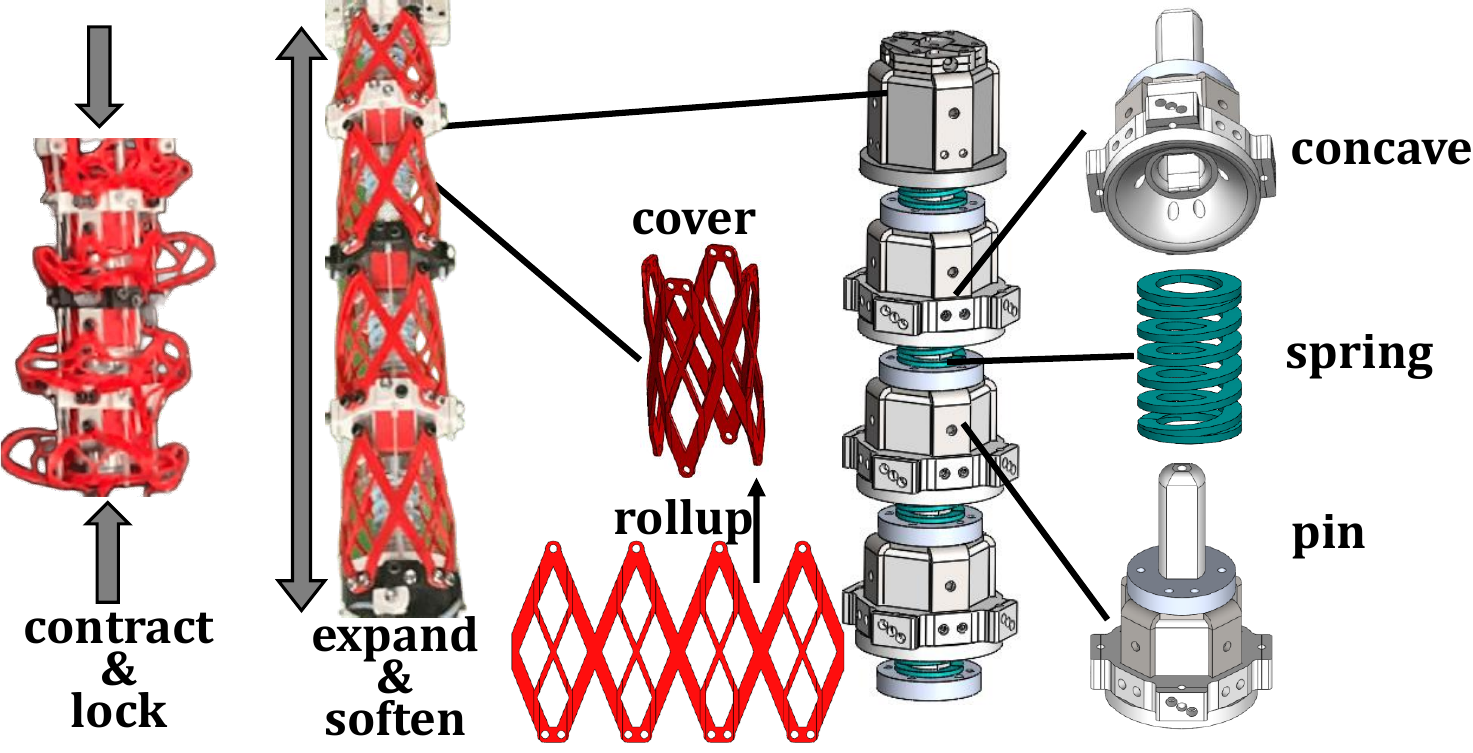}
        \caption{Overview of the Variable-Stiffness Contract Link (VSC-Link). By applying expansion and contraction forces with a spring and a wire, respectively, to the interlocking concave and convex parts, expansion/contraction and stiffness switching are performed. Four sets of concave, convex, spring, and cover are connected in series.}
    \vspace{-2.0ex}
    \label{fig:VSC-Link}
\end{figure}

To satisfy all of the above requirements, we adopted a structure that locks into a rigid state when contracted beyond a certain point. The specific structure is shown in \figref{fig:VSC-Link}. Convex and concave parts are connected in series with a spring in between, and are driven by the extension force of the spring (maximum 244 N) and the contraction force of the wire. When the contraction exceeds a certain level, the pin of the convex part is guided into the back of the concave part, creating a structure that locks against bending and twisting. In addition, it is constrained by a flexible cover to prevent extension beyond the natural length of the spring and excessive twisting.

\subsubsection{\titlecap{AGI-EE}}
\label{subsubsec:AGI-EE}
\begin{figure}
    \centering
    \includegraphics[width=0.5\linewidth]{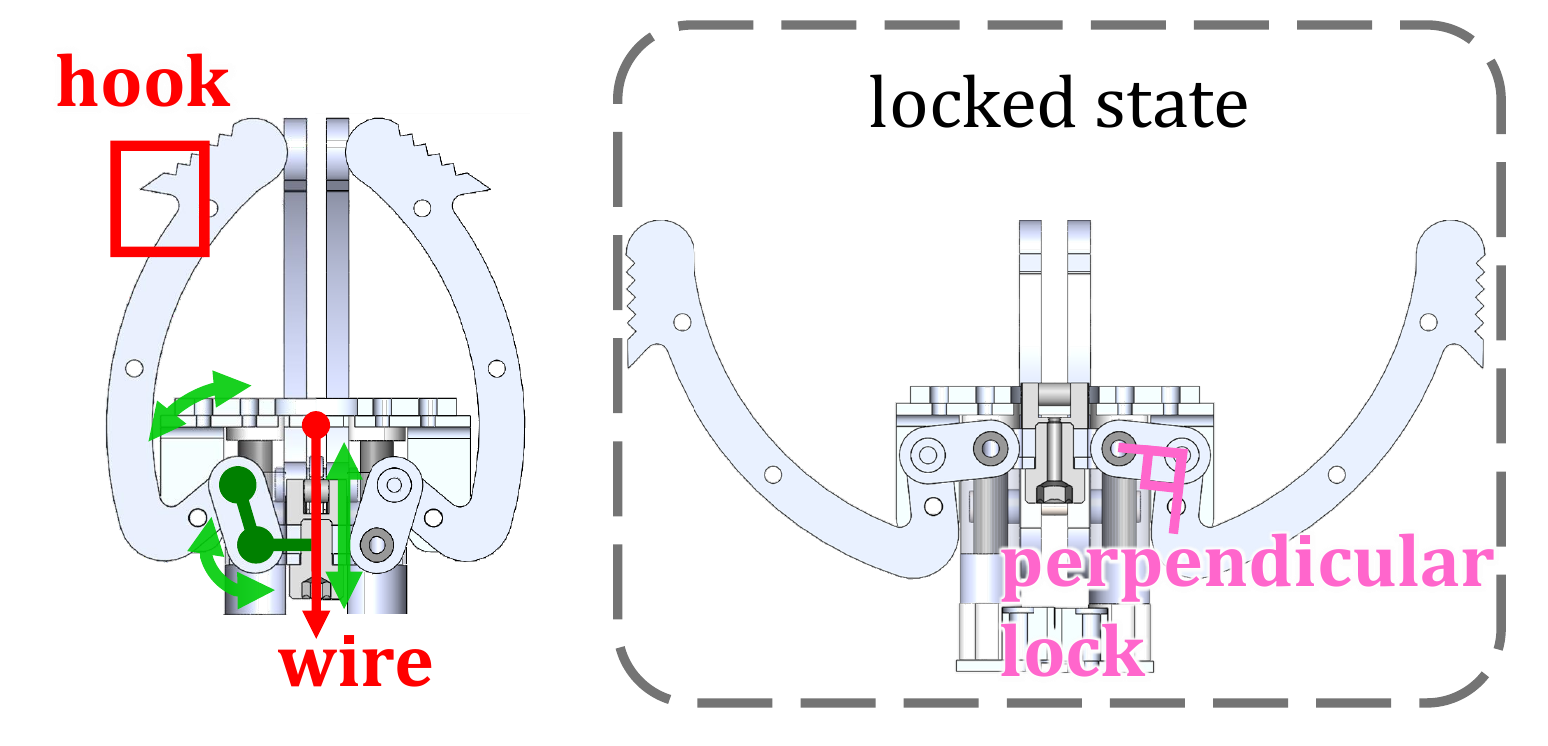}
    \vspace{-1.0ex}
        \caption{The AGI-EE has a structure that converts the linear motion of the wire and spring into the rotation of the fingers. When fully opened, it can be locked against external forces and used as an anchor, and peristaltic motion is possible in combination with the expansion and contraction of the link.}
    \vspace{-2.0ex}
    \label{fig:AGI-EE}
\end{figure}
\begin{figure}
    \centering
    \includegraphics[width=0.5\linewidth]{figs/gripper_torque.pdf}
        \caption{AGI-EE opening and closing torque. In the fully opened state, the opening torque diverges and the lock is engaged, while in most other angle ranges, the torque in both directions is stable.}
    \vspace{-2.0ex}
    \label{fig:AGI-EE_torque}
\end{figure}
In addition to grasping objects, the AGI-EE also functions as an anchor that catches on the environment during peristaltic motion. The structure is shown in \figref{fig:AGI-EE}. It is a link mechanism that converts the linear motion of the wire into the rotation of the fingers, with closing performed by wire tension (450 N) and opening by a spring that counteracts the wire (maximum extension force 149 N). There are hooks on the outside of the fingers for catching on the environment. In the fully opened state, the link becomes perpendicular and locks against external forces, so it does not close even when a strong force is applied by catching on the environment, and can maintain the anchor state. The finger torque is shown in \figref{fig:AGI-EE_torque}. While it is locked in the fully opened state, it exerts a stable torque of -2 to 4 N$\cdot$m in most angle ranges. Therefore, it can be seen that it can function as both an anchor and a gripper.

\section{\titlecap{Control Architecture}}\label{sec:control}
This section describes the control of REWW-ARM.
\subsection{\titlecap{AGI-EE} and \titlecap{distal VSC-Link control}}
The AGI-EE and the distal VSC-Link each correspond to one wire. They are controlled by applying discrete tensions of 450 N and 300 N, respectively.

\subsection{\titlecap{Joints} and \titlecap{proximal VSC-Link control}}
\begin{figure}
    \centering
    \includegraphics[width=0.8\columnwidth]{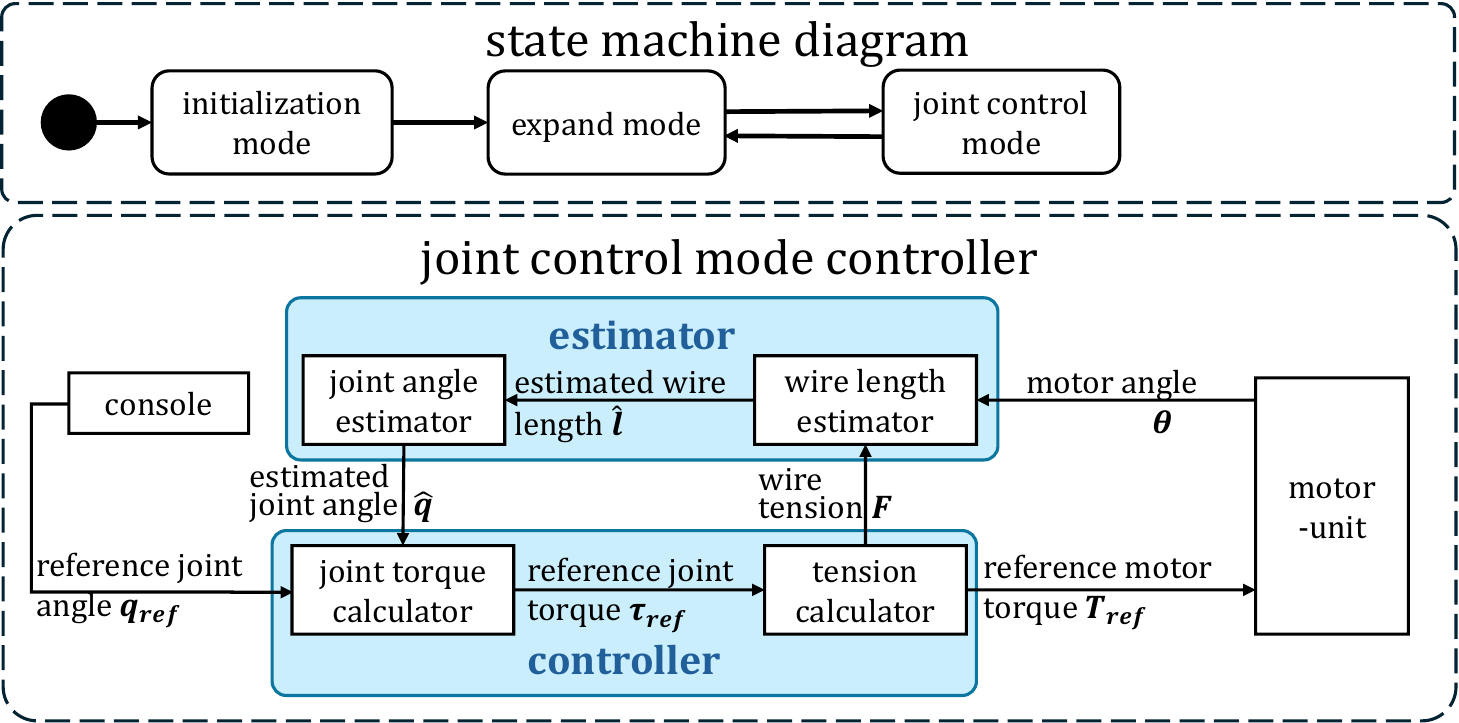}
        \caption{Overview of REWW-ARM control. After initialization such as offset acquisition after startup, the expansion mode and joint control mode are switched by console input. The joint control mode controller consists of two parts: an estimator that estimates the joint angle through the wires, and a follower that works to make the estimated joint angle follow the target joint angle. The estimator estimates the wire length from the wire tension and motor angle and estimates the joint angle by the weighted least squares method. The follower calculates the joint torque by PID control from the difference between the estimated joint angle and the target joint angle, then calculates the tension for each wire with constraints, converts it to motor torque, and commands it.}
    \vspace{-2.0ex}
    \label{fig:control_summary}
\end{figure}

As shown in \figref{fig:mobile_robot_part}, the three joints of the shoulder and elbow are driven by four wires. Furthermore, all four wires pass through the proximal VSC-Link so as to contract it. In summary, the structure is such that the three joints are driven by the difference in tension of the four wires, and the VSC-Link contracts due to the sum of the tensions. Therefore, REWW-ARM has the following three modes as shown in \figref{fig:control_summary}.
\begin{itemize}
    \item \textbf{initialization mode}: A mode to acquire the motor angle offset for a few seconds after startup. By applying a weak tension with the joint angles manually set to 0, a value with little influence from elongation or slack is obtained.
    \item \textbf{expansion mode}: A mode in which the tension of the four wires is set to 0, and the proximal VSC-Link is fully extended.
    \item \textbf{joint control mode}: A mode in which the sum of the tensions of the four wires is brought close to a specific value (340 N), and the joint angles are controlled by the difference in tension while the proximal VSC-Link is fully contracted.
\end{itemize}

The controller in joint control mode consists of an \textbf{estimator} and a \textbf{follower}, as shown in \figref{fig:control_summary}. The estimator estimates the joint angle $\bm{q}$ from the wire tension and displacement, and the follower calculates the wire tension such that the estimated joint angle $\hat{\bm{q}}$ follows the target joint angle $\bm{q}_\textrm{ref}$.

\subsubsection{\titlecap{Estimator}}
The estimator estimates the estimated joint angle $\hat{\bm{q}}$ according to the following procedure.
\begin{enumerate}
    \item[i)] \textbf{Wire length estimation}: 
    Let $\bm{\theta}$ be the motor angle minus the offset obtained in initialization mode, $r_\textrm{pulley}$ be the winding winch radius, $k$ be the wire elongation per unit tension, $\bm{F}$ be the wire tension, and $\bm{q}$ be the joint angle of the distal mobile robot. At this time, the estimated value $\hat{\bm{l}}$ of the wire displacement $\bm{l}$ based on the time when $\bm{q}=\bm{0}$ is as shown in \equref{eq:estim_wirelen}.
    \begin{equation}
        \hat{\bm{l}} = r_\textrm{pulley}\bm{\theta} - k \bm{F}
        \label{eq:estim_wirelen}
    \end{equation}
    \item[ii)] \textbf{Wire length centering}: 
    The proximal VSC-Link expands and contracts slightly even in the locked state, and its effect is almost uniform for all wire lengths. Here, since the tendon Jacobian $\bm{G}$ is a constant regardless of $\bm{q}$, and $\bm{l}$ is defined such that $\bm{l}=\bm{0}$ when $\bm{q}=\bm{0}$, \equref{eq:tendon_len_eq} is derived from \equref{eq:tendon_len_eq_der}.
    \begin{equation}
        \bm{l} = \bm{G} \bm{q}
        \label{eq:tendon_len_eq}
    \end{equation} At this time, from the value of $\bm{G}$ shown in \equref{eq:tendon_Jacobian_val}, the sum of the values of the elements of $\bm{l}$ is 0 for any $\bm{q}$. Therefore, the effect of expansion and contraction can be canceled by subtracting the average value $\bar{\hat{\bm{l}}}$ from $\hat{\bm{l}}$ to obtain the centered wire length $\tilde{\bm{l}}$.

    \item[iii)] \textbf{Angle estimation by weighted least squares}: 
    Since there are 4 wires passing through 3 joints, $\bm{q}$ is estimated from $\tilde{\bm{l}}$ by the least squares method. Here, since the tension distribution within the RWTM for each wire is not necessarily uniform, the aforementioned estimated wire length $\tilde{\bm{l}}$ is not necessarily accurate. Its reliability is low at tensions below a certain level because slack is expected, and it decreases as the tension rises above a certain level due to the effect of elongation. To utilize this reliability in joint angle estimation, the weighted least squares method is used. For each wire, the weight $w$ is determined for the tension $F$ as shown in \equref{eq:wire_weight}.
    \begin{equation}
        w = 
            \begin{cases}
            0.5 & \text{if } 0 < F < 10 \\
            \exp(-\frac{F-10}{1000}) & \text{if } 10 \leq F
            \end{cases}
        \label{eq:wire_weight}
    \end{equation}

    When the matrix with these $w$ values arranged diagonally for each wire is denoted as $\bm{W}$, the estimated joint angle $\hat{\bm{q}}$ can be obtained by solving the optimization problem \equref{eq:wls}.
    \begin{equation}
        \min_{\bm{q}} \left( 
             (\tilde{\bm{l}}-\bm{G}\bm{q})^T \bm{W} (\tilde{\bm{l}}-\bm{G}\bm{q}) 
             \right)
        \label{eq:wls}
    \end{equation}

\end{enumerate}
\subsubsection{\titlecap{Follower}}
The follower calculates the command motor torque $\bm{T}_\textrm{ref}$ according to the following procedure.

\begin{enumerate}
    \item[i)] \textbf{Joint torque calculation}: 
    The target joint torque $\bm{\tau}_\textrm{ref}$ is calculated by the PID method from the difference between the estimated joint angle $\hat{\bm{q}}$ and the target joint angle $\bm{q}_\textrm{ref}$.

    \item[ii)] \textbf{Tension calculation}: 
    The relationship between the tension $\bm{F}$ and the joint torque $\bm{\tau}$ is as shown in \equref{eq:tendon_torque_eq}. Since there are 4 wires driving 3 joints, there are infinitely many solutions $\bm{F}$ that satisfy \equref{eq:tendon_torque_eq} for a desired joint torque $\bm{\tau}_\textrm{ref}$. Therefore, the following constraints can be given: 1) satisfy the minimum tension $F_\textrm{min}=$10 N to prevent the wires from slacking and the maximum tension $F_\textrm{max}=$500 N to prevent the motors from overheating; 2) ensure that the sum of tensions is as close as possible to the target total tension $F_\textrm{contract}=$340 N to maintain the maximum contraction state of the proximal VSC-Link. Note that this target total tension was set with a margin from the maximum extension force of 244 N by the spring. Formulating these, the target tension $\bm{F}_\textrm{ref}$ is the solution to the optimization problem \equref{eq:torque_to_tension_opt}.
    \begin{equation}
    \begin{aligned}  
    &\min  \qquad \left( \left(\sum_{i=1}^{4}F_i\right) - F_\textrm{contract} \right)^2
    \\ 
    &\text{s.t.} 	 \left(
    \begin{aligned}
        &\bm{\tau} = - \bm{G}^T \bm{F}\\
        &{F_\textrm{min}} < F_i < F_\textrm{max} &  (1 \leq i \leq 4)
    \end{aligned}
    \right)
    \end{aligned}
    \label{eq:torque_to_tension_opt}
    \end{equation}

    Note that the torque range mentioned in \secref{sec:design} generally takes these constraints into account. The tension $\bm{F}$ used here is an approximation based on the tension measured at the motor side. Accurately estimating the true tension at the distal end is challenging due to complex friction and hysteresis phenomena within the RWTM. While this study uses a simplified model, developing more sophisticated friction models is a key area for future work to improve control accuracy.
    \item[iii)] \textbf{Motor torque calculation}: 
    The target motor torque $\bm{T}_\textrm{ref}$ is converted by multiplying the previously obtained target tension $\bm{F}_\textrm{ref}$ by $r_\textrm{pulley}$.
\end{enumerate}

\subsection{\titlecap{peristalsis control}}
\label{subsec:proceed_control}
\begin{figure}%
    \centering
    \includegraphics[width=0.5\linewidth]{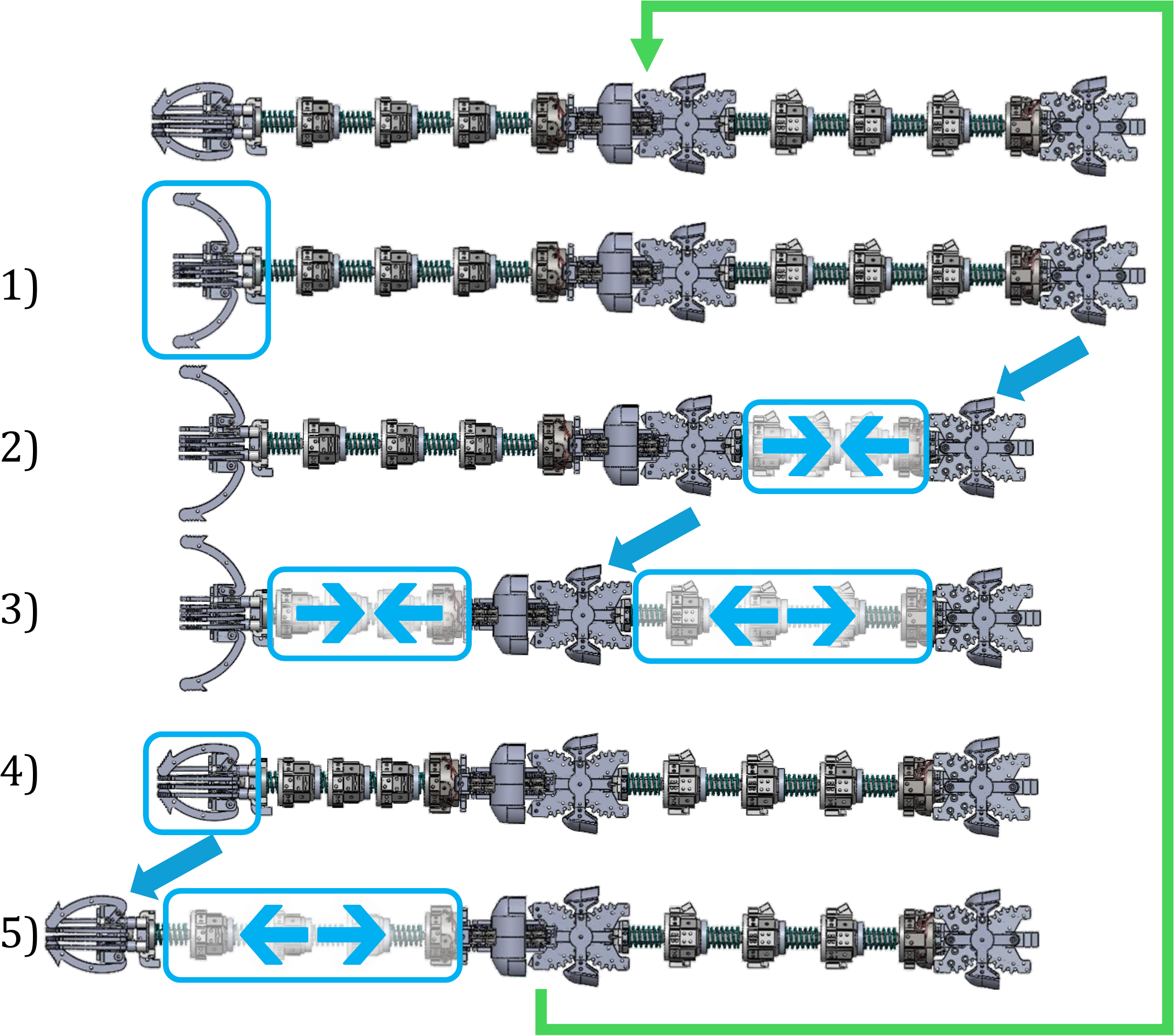}
        \caption{Procedure for peristaltic motion. By linking the opening and closing of the AGI-EE with the expansion and contraction of the VSC-Link, the shoulder, elbow, and AGI-EE are moved forward in order.}
    \vspace{-2.0ex}
    \label{fig:proceed_seq}
\end{figure}

The distal mobile robot can move by peristalsis as shown in \figref{fig:proceed_seq} by repeating the following operations.
\begin{enumerate}
    \item Open the AGI-EE and hook it to the environment.
    \item Contract the proximal VSC-Link and move the shoulder forward.
    \item While contracting the distal VSC-Link, extend the proximal VSC-Link and move the elbow forward.
    \item Close the AGI-EE and release the hook from the environment.
    \item Extend the distal VSC-Link and move the AGI-EE forward as well.
\end{enumerate}

\section{\titlecap{Experiments and Results}}\label{sec:experiments}
\subsection{\titlecap{grasping experiment}}
\begin{figure}
    \centering
    \includegraphics[width=0.7\columnwidth]{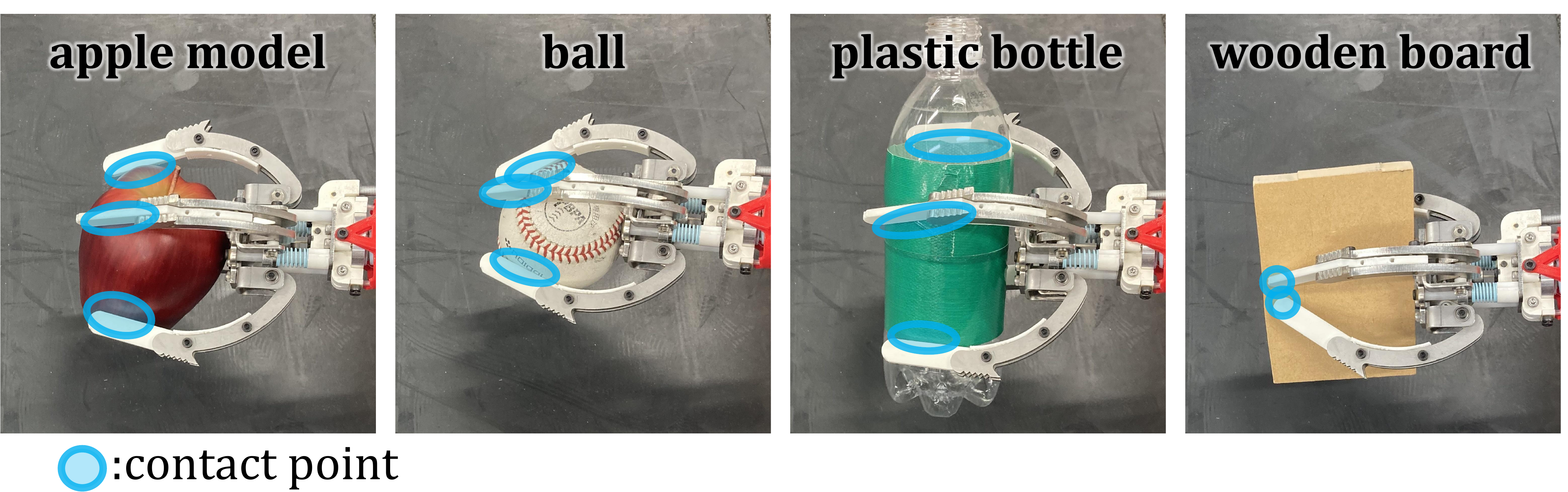}
        \caption{Grasping experiment. We confirmed that objects such as an apple model, a ball, a plastic bottle, and a wooden board could be grasped and held in the air.}
    \vspace{-2.0ex}
    \label{fig:grasp_experiment}
\end{figure}

As shown in \figref{fig:grasp_experiment}, we confirmed that various objects can be held in the air using the AGI-EE. The AGI-EE has four arc-shaped fingers attached axisymmetrically. Therefore, small spherical objects such as an apple model and a ball, and cylindrical objects such as a plastic bottle could be grasped by wrapping around them, and they could be held stably even when shaken by an external force. On the other hand, when grasping a thin, wide wooden board, the AGI-EE was almost in a closed state, and the contact points between the fingers and the object were concentrated in a small area. Therefore, when the AGI-EE was shaken a little, the wooden board rotated within the AGI-EE, which was somewhat unstable.
\subsection{\titlecap{Preliminary Transmission Characteristics Evaluation}}
\label{subsec:RWTM_single_exp}
\begin{figure}
    \centering
    \includegraphics[width=0.7\columnwidth]{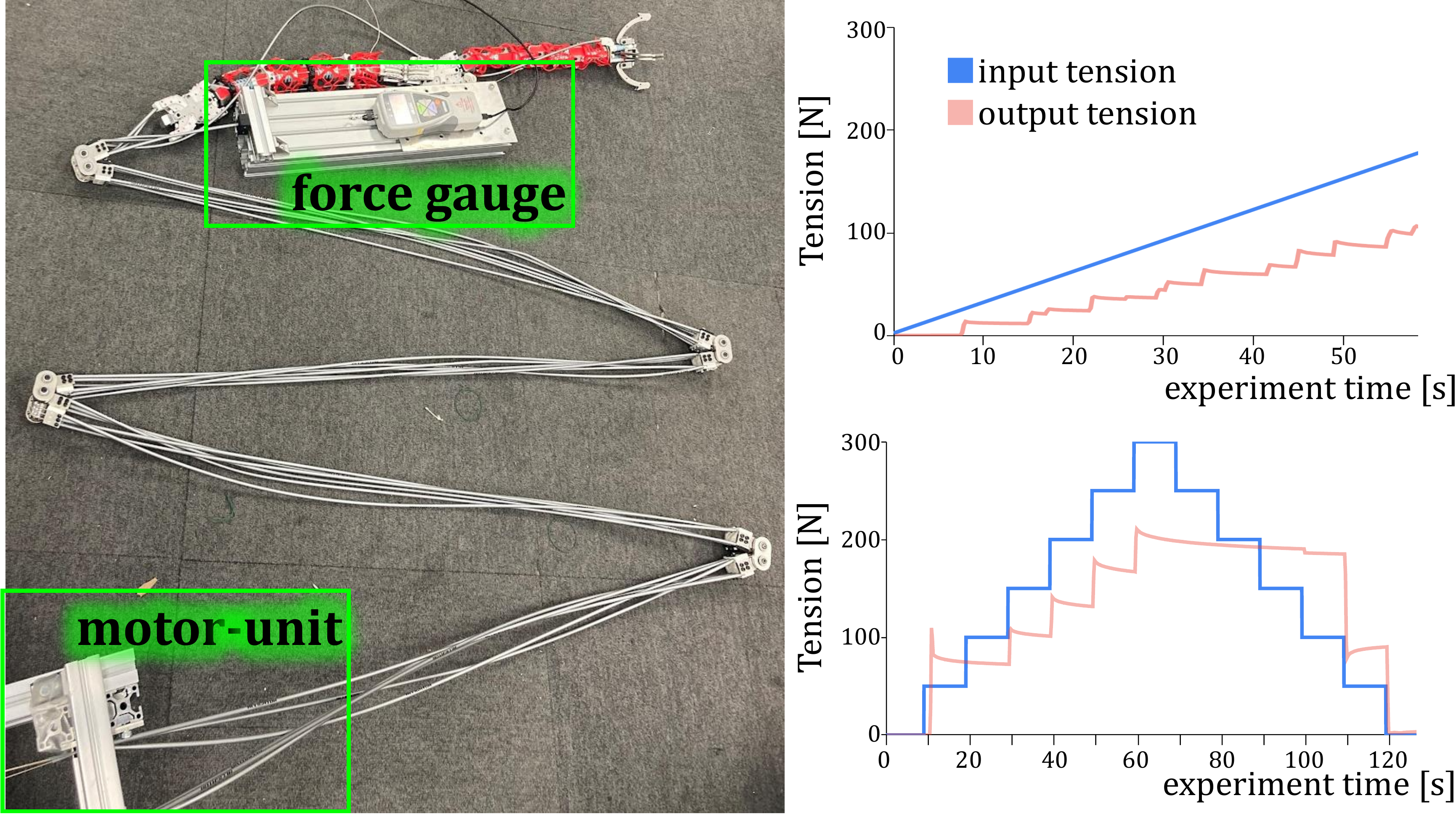}
        \caption{Transmission characteristics measurement experiment with a single wire RWTM. The relationship between the command tension at the motor unit and the output tension at the distal mobile robot side was investigated. Hysteresis due to friction was confirmed.}
    \vspace{-2.0ex}
    \label{fig:RWTM_single_exp}
\end{figure}

We conducted a preliminary experiment to investigate the basic transmission characteristics of the RWTM. By connecting a force gauge (IMADA ZTS-500N) to the output end of a single wire path and commanding tension from the motor, we observed the relationship shown in \figref{fig:RWTM_single_exp}. When the command tension was increased linearly, the output tension exhibited step-like increases. When the tension was changed in steps, the output pattern more closely followed the input, suggesting that discrete control might improve responsiveness. A clear hysteresis due to friction was also confirmed. This preliminary result motivated a more detailed investigation into the transmission efficiency, as described in the next section, and informed the design of the discrete torque controller evaluated in \secref{subsec:controller_exp}.

\subsection{\titlecap{Detailed Transmission Characteristics Evaluation}}
\label{subsec:detail_eval}
To quantitatively evaluate the advantages of the RWTM design, we conducted a detailed experiment comparing the transmission efficiency of the RWTM with that of a pure TSM and a series of decoupled joints alone. 

\begin{figure}
    \centering
    \includegraphics[width=1.0\linewidth]{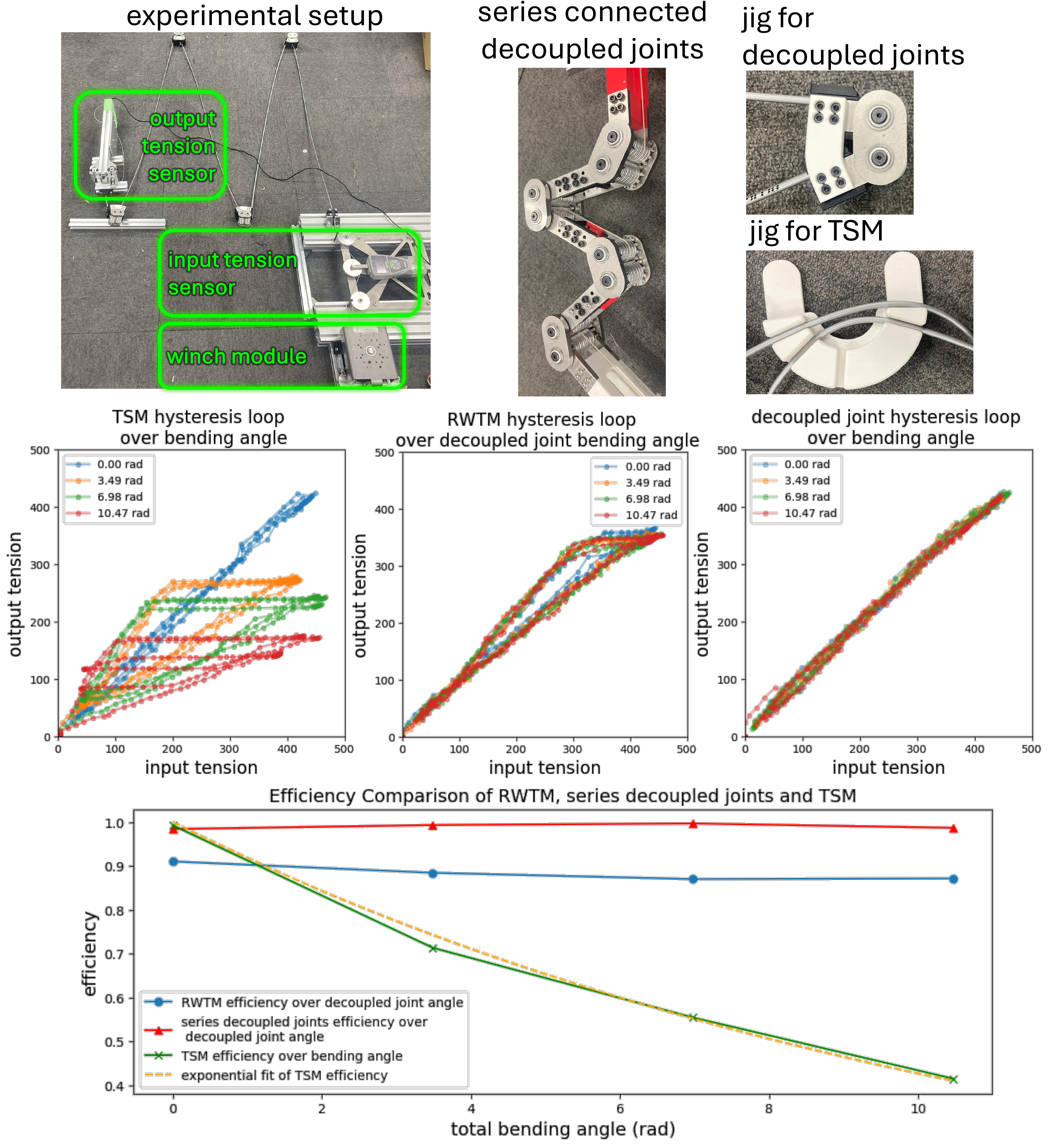}
        \caption{Detailed transmission characteristics experiment. (Top) Experimental setup for measuring input and output tension for different transmission line configurations (decoupled joints only, TSM only, and full RWTM) at various bending angles. (Middle) Hysteresis loops showing the relationship between input and output tension for each configuration and angle. (Bottom) Transmission efficiency plotted against bending angle, showing a comparison among experimental data of each configuration and the theoretical capstan friction model for the TSM.}
    \vspace{-2.0ex}
    \label{fig:rwtm_eff_detail_exp}
\end{figure}

\textbf{Setup:} As shown in \figref{fig:rwtm_eff_detail_exp}, we measured the input and output tensions of three different transmission lines: (1) four decoupled joints in series, (2) a 1-meter-long TSM, and (3) the full RWTM (four decoupled joints and TSM sections). For each setup, we applied a tension profile that ramped from 10 N to 400 N and back down, while bending the joints and sheath at various cumulative angles (0, 200, 400, and 600 degrees) using a custom jig to ensure accuracy.

\textbf{Results:} The resulting hysteresis loops are shown in \figref{fig:rwtm_eff_detail_exp}. From these, we calculated the average transmission efficiency (ratio of output to input tension during the tension-increasing phase). The results, summarized in \tabref{tab:eff_comp} and \figref{fig:rwtm_eff_detail_exp}, clearly show that the decoupled joints maintain a very high and stable efficiency ($\geq$98\%) regardless of bending angle. In contrast, the TSM's efficiency drops sharply with increased bending, as predicted by the capstan friction model presented in \equref{eq:capstan_friction}. By fitting this model to the data, we estimated the friction coefficient $\mu$ for our TSM to be approximately \textbf{0.085}. The RWTM, combining both TSM and decoupled joints, demonstrates significantly higher and more stable efficiency than the TSM alone across the tested angles. The average efficiency of the RWTM was 0.884 (SD=0.016), whereas the TSM's was 0.669 (SD=0.215). The RWTM's efficiency surpassed that of the TSM over 85\% of the bending range achievable by its decoupled joints.

\begin{table}[htbp]
    \centering
    \caption{Comparison of transmission efficiency among different transmission line configurations.}
    \begin{tabular}{|p{0.29\columnwidth}|p{0.28\columnwidth}|p{0.28\columnwidth}|} 
        \hline 
        & Average Efficiency & Std. Dev. of Efficiency \\ \hline
        RWTM & \textbf{0.884} & \textbf{0.016} \\ \hline
        series of decoupled joints & 0.991 & 0.005 \\ \hline
        TSM only & 0.669 & 0.215 \\ \hline
    \end{tabular}
    \vspace{-2.0ex}
    \label{tab:eff_comp}
\end{table}

\subsection{\titlecap{Controller performance evaluation experiment}}
\label{subsec:controller_exp}
In the results of \subsecref{subsec:RWTM_single_exp}, when the input tension was changed in steps, the output tension followed a pattern closer to the input tension than when it was changed linearly. From this, we considered that the controller performance could be improved by changing the tension discretely, and devised a control method that discretizes the target joint torque calculated by PID every 5 N$\cdot$ m (hereinafter referred to as discrete torque control). Then, we evaluated the controller performance with and without discretization.
The evaluation experiment environment and system are shown in \figref{fig:anglefollow_exp}. A cube with AR markers was fixed to the AGI-EE, and the position and orientation of the marker box were acquired from a camera. The actual joint angles were obtained by solving inverse kinematics (IK) for that position and orientation. For the IK solver, we used Powell's method, a gradient-free sequential optimization algorithm available in "scipy.optimize.minimize". This method is robust for non-linear and non-differentiable cost functions and is less prone to divergence at singularities. By using the solution from the previous time step as the initial guess for the current one, the computation is fast enough to run at the 10 Hz control loop frequency. Then, the time-series data was compared with the time-series data of the target joint angles and the estimated joint angles from the wires to evaluate the performance of the follower and the estimator. Although there are actually three joints, the VSC-Link shows slight expansion and contraction even in the locked state, so it was treated as a virtual fourth joint when solving IK.
Experiments were conducted by giving random joint angle commands from -1.3 to 1.3 rad to each joint under the conditions with and without discretization, and the results are shown in \figref{fig:anglefollow_exp}.

We first evaluated this result using normalized cross-correlation (NCC). Considering the time delay $d$ of image processing, etc., the correlation between any time series $y, z$ was calculated as in \equref{eq:std_corr}.
\begin{equation}
    \begin{aligned}
    &\max_{d} \left( \frac{\sum_{t} (y(t)-\bar{y})(z(t-d)-\bar{z})}{\sqrt{\sum_{t} (y(t)-\bar{y})^2}\sqrt{\sum_{t} (z(t-d)-\bar{z})^2}} \right)
    \end{aligned}
    \label{eq:std_corr}
\end{equation}

In addition, we also calculated the MSE error. When these were calculated for each joint and averaged, the results were as shown in \tabref{tab:corr}. Here, the follower performance was calculated between the target joint angle and the estimated joint angle, the estimator performance was calculated between the estimated joint angle and the actual measured joint angle, and the integrated performance was calculated between the target joint angle and the actual measured joint angle. Torque discretization was slightly superior only in the estimator NCC. Overall, the non-discretized controller showed better integrated performance.

\begin{figure}
    \centering
    \includegraphics[width=0.7\linewidth]{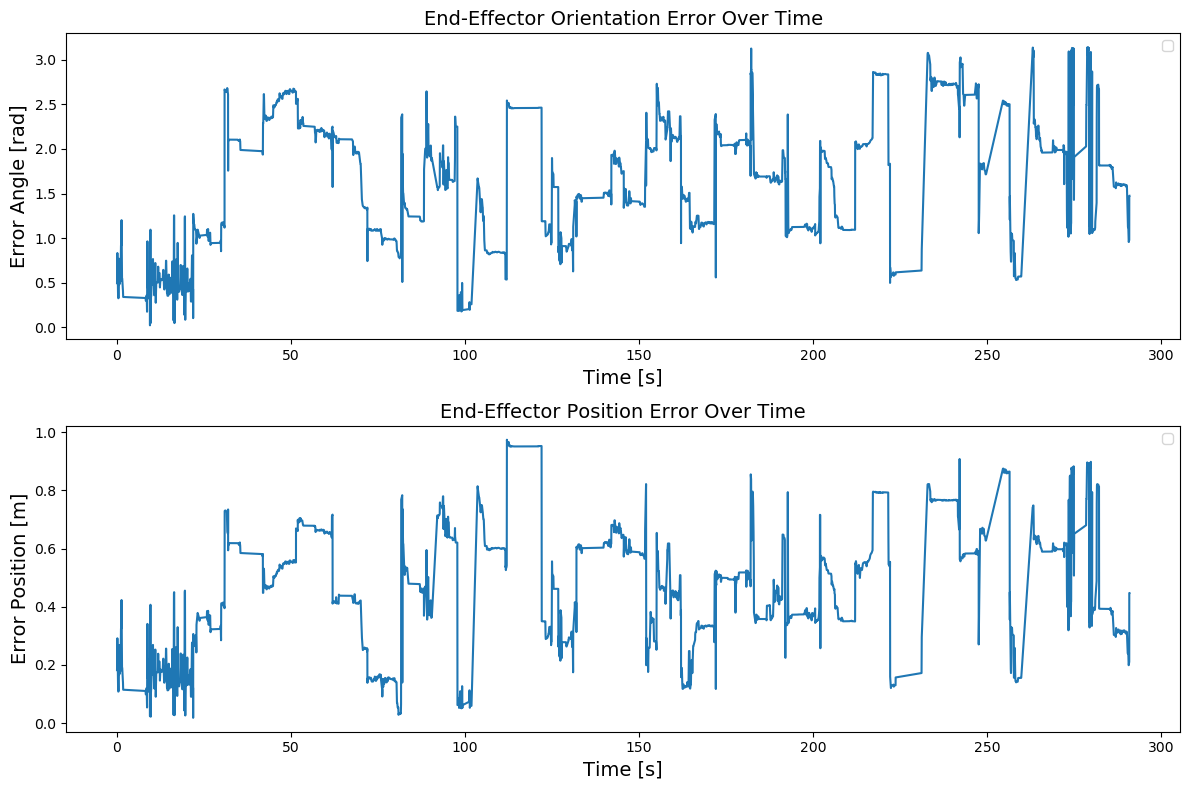}
        \caption{End-effector position and orientation error over time for the controller performance evaluation experiment without torque discretization.}
    \vspace{-2.0ex}
    \label{fig:EE-error}
\end{figure}

\begin{figure}
    \centering
    \includegraphics[width=1.0\linewidth]{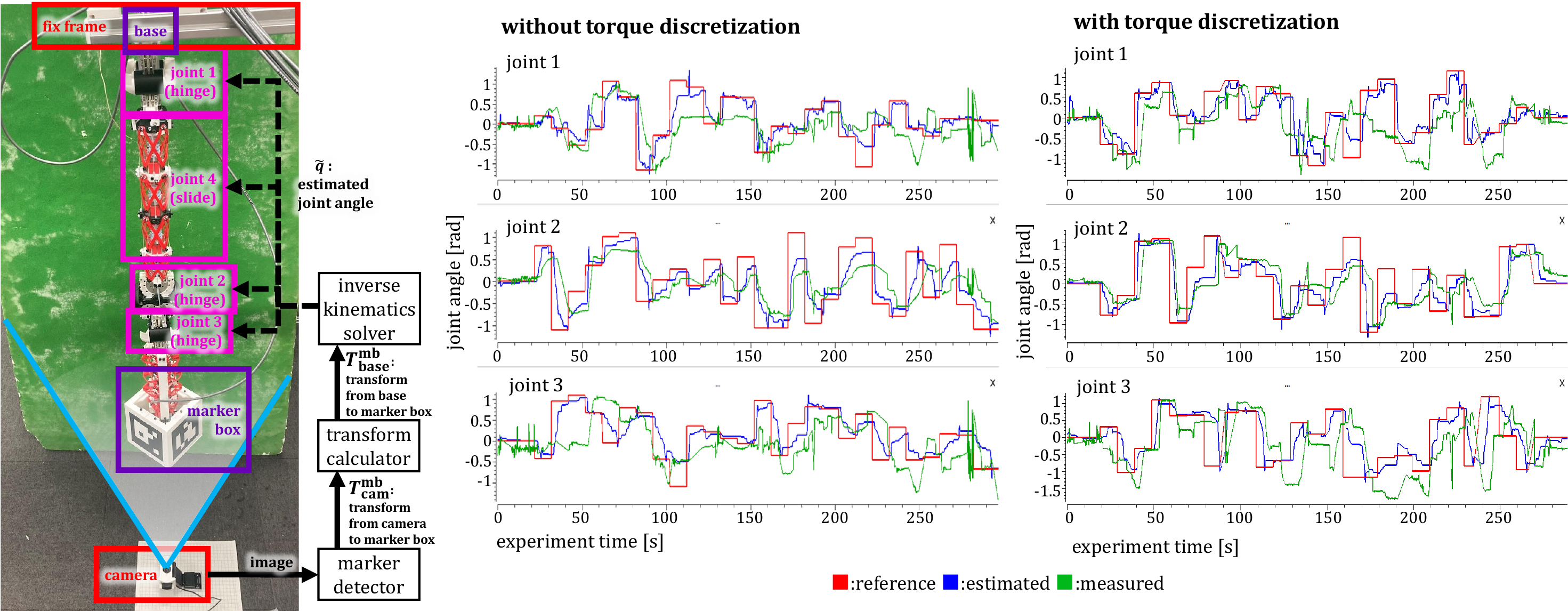}
        \caption{System configuration and results of the controller performance evaluation experiment. A marker box was fixed to the AGI-EE, and its position and orientation were read from a camera to solve IK and measure the actual joint angles. The results were compared with the target joint angles and the estimated joint angles within the controller.}
    \vspace{-2.0ex}
    \label{fig:anglefollow_exp}
\end{figure}

Furthermore, we evaluated the end-effector pose error for the non-discretized case, as shown in \figref{fig:EE-error}. The position error is the Euclidean distance between the target and actual positions of the AGI-EE. The orientation error is the angle of the axis-angle representation of the relative rotation between the target and actual orientations. The average position error was 0.42 m, and the average orientation error was 1.52 rad. These errors are primarily attributed to the unmodeled friction and hysteresis in the RWTM, which is a target for future improvement.

\begin{table}[htbp]
    \centering
    \caption{The performance of the estimator and follower was evaluated by NCC and MSE, respectively, when torque discretization was applied and when it was not.}
    \begin{tabular}{|p{0.29\columnwidth}|p{0.17\columnwidth}|p{0.17\columnwidth}|p{0.17\columnwidth}|} 
        \hline 
        &follower NCC & estimator NCC & integrated NCC \\ \hline
        with torque\par discretization & 0.8854 & \textbf{0.7675} & 0.599 \\ \hline
        without torque\par discretization & 0.9058 & 0.6840 & 0.596\\ \hline
    \end{tabular}
    \vspace{5pt} \\
    \begin{tabular}{|p{0.29\columnwidth}|p{0.17\columnwidth}|p{0.17\columnwidth}|p{0.17\columnwidth}|} 
        \hline 
        &follower MSE [rad$^2$] & estimator MSE [rad$^2$] & integrated MSE [rad$^2$]\\ \hline
        with torque\par discretization & 0.1178 & 0.2001 & 0.3837 \\ \hline
        without torque\par discretization & 0.0779 & 0.1714 & 0.2910 \\ \hline
    \end{tabular}
    \vspace{-2.0ex}
    \label{tab:corr}
\end{table}

\subsection{\titlecap{locomotion experiment}}
\begin{figure}
    \centering
    \includegraphics[width=1.0\linewidth]{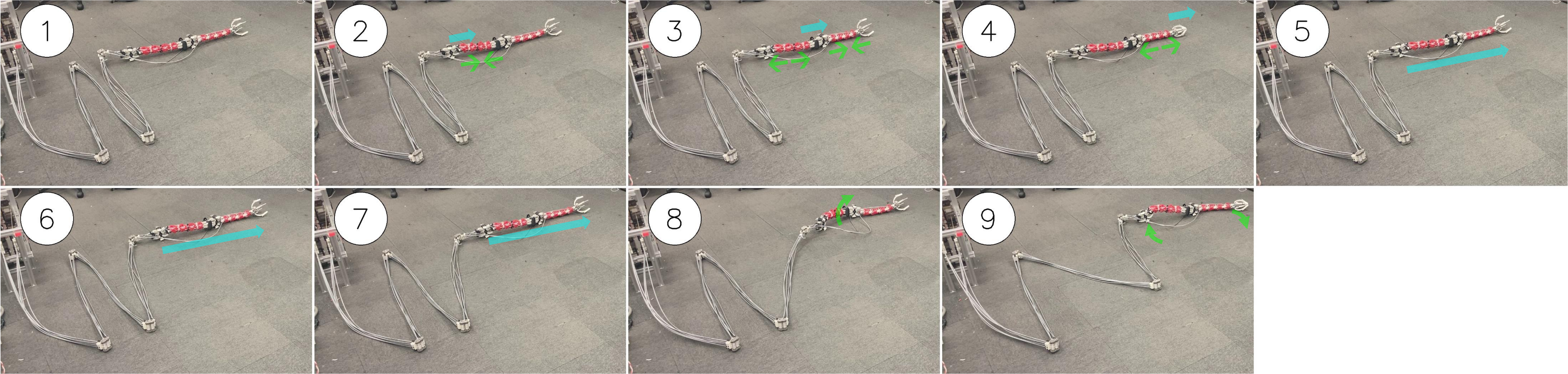}
        \caption{locomotion experiment. We performed peristaltic motion and rotational motion by coordinating the VSC-Link, GCD-Joint, and AGI-EE. After advancing about 0.45 m, we made a direction change of about 15 degrees.}
    \vspace{-2.0ex}
    \label{fig:proceed_exp}
\end{figure}

We verified the straight-line locomotion ability and rotational locomotion ability. The experiment is shown in \figref{fig:proceed_exp}. First, we performed the peristaltic motion described in \subsecref{subsec:proceed_control} 10 times in 110 seconds and advanced 0.45 m. After that, we lifted the upper arm using joint2, which has a pitch degree of freedom, and then rotated the upper arm to the left by moving joint3, which has a yaw degree of freedom, clockwise. After that, we lowered the upper arm to the ground and then moved joint3 counterclockwise, which rotated the forearm to the right. This caused the entire distal mobile robot to rotate to the right by about 15 degrees.

\subsection{\titlecap{underwater experiment}}
\begin{figure}
    \centering
    \includegraphics[width=1.0\linewidth]{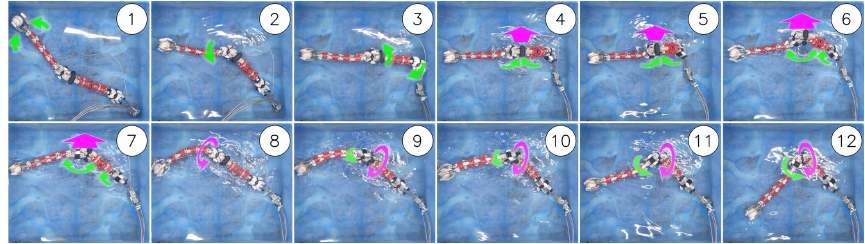}
        \caption{We confirmed the operation underwater. With the entire robot and part of the RWTM submerged, we performed grasping, joint control, and parallel translation and roll rotation. No major effects on operation due to submersion were observed.}
    \vspace{-2.0ex}
    \label{fig:water_exp}
\end{figure}

We verified the operational capability underwater. As shown in \figref{fig:water_exp}, the entire distal mobile robot and one of the decoupled joints were completely submerged in water at a depth of about 0.2 m. In this state, we performed the following movements and confirmed that object grasping and position/posture control can be performed even underwater.
\begin{enumerate}
    \item \textbf{Grasping}: In \figref{fig:water_exp}\circletext{1}, we grasped a baseball with a diameter of 67 mm.
    \item \textbf{Yaw rotation}: From \figref{fig:water_exp}\circletext{2} to \circletext{3}, we drove joint3 counterclockwise to move the forearm downward in the figure, and then used joint2 to lift the upper arm slightly. Then, we drove joint3 clockwise to rotate the entire body to be horizontal on the screen. This caused the entire distal mobile robot to rotate counterclockwise by about 45\textdegree in the figure.
    \item \textbf{Parallel translation}: In \circletext{4} and \circletext{5}, we bent joint2 to raise the elbow above the water, and then from \circletext{6} to \circletext{8}, we returned joint1 to a straight line while returning joint2 to a straight line. This caused it to fall upward in the figure due to gravity, and the entire body moved slightly upward in parallel.
    \item \textbf{Roll rotation}: From \circletext{8} to \circletext{12}, we drove joint2 to raise the forearm in the lower left direction of the screen, thereby using the gravity on the forearm to roll forward by about 90 degrees.

\end{enumerate}

\section{\titlecap{Discussion}}\label{sec:discussion}

\subsection{\titlecap{RWTM Design Trade-offs and Limitations}}

The design of the RWTM involves a trade-off between the number of decoupled joints ($N$) and the total length of the TSM ($L_\textrm{tsm}$) for a given total length ($L_\textrm{total}$). The optimal configuration and maximum total length depends on four main factors below.

\begin{figure}
    \centering
    \includegraphics[width=0.6\linewidth]{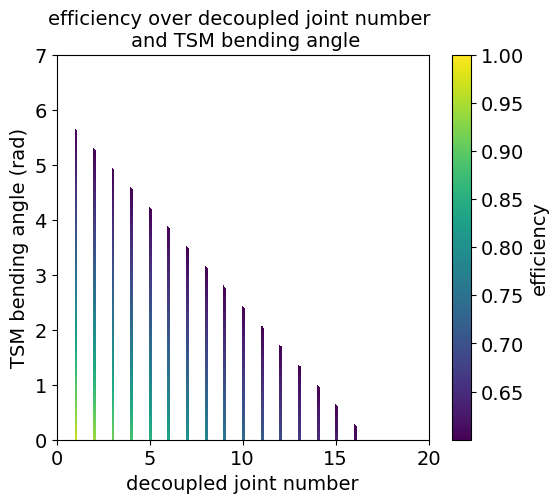}
        \caption{RWTM efficiency constraint. The colored region shows the combinations of the number of decoupled joints ($N$) and the TSM bending angle ($\theta_\textrm{tsm}$) where the transmission efficiency remains above 60\%.}
    \vspace{-2.0ex}
    \label{fig:rwtm_eff_over_dec_num_and_tsm_angle}
\end{figure}

\begin{itemize}
    \item \textbf{Weight Constraint:} The total mass of the RWTM ($M_\textrm{total}$) increases with both $N$ and $L_\textrm{total}$. Based on the decoupled joint mass $M_\textrm{dec}=0.34$ kg, decoupled joint length $L_\textrm{dec}=0.096$ m, TSM bundle weight per length $M_\textrm{tsm}=0.36$ kg/m, the total mass is $M_\textrm{total} \approx 0.305N + 0.36L_\textrm{total}$. The traction force of the distal mobile robot must overcome the friction of the RWTM, imposing a weight constraint. For a distal robot with mass $M_\textrm{tip}$ and friction coefficient against the ground $\mu_\textrm{tip}$, its maximum traction force is estimated to be $M_\textrm{tip}\mu_\textrm{tip}g$ for most major locomotion mechanism such as wheels, legs, and peristalsis mechanism. Therefore, the condition is approximately $M_\textrm{tip}\mu_\textrm{tip} \geq (0.305N+0.36L_\textrm{total})\mu $, where $\mu_\textrm{rwtm}$ is the friction coefficient between the RWTM and the ground.

    \item \textbf{Efficiency Constraint:} The overall efficiency depends on $N$ and the bending angle of the TSM, $\theta_\textrm{tsm}$. Using the efficiency per decoupled joint ($\eta_\textrm{dec} = 0.97$) with connecting friction taken into account and the TSM friction coefficient ($\mu_\textrm{tsm} = 0.085$) calculated from \secref{subsec:detail_eval}, the total efficiency is $\eta_\textrm{total} = (\eta_\textrm{dec})^N \exp(-\mu_\textrm{tsm}\theta_\textrm{tsm})$. To maintain an efficiency above a certain threshold (e.g., 60\%), there is a constraint on the allowable combination of $N$ and $\theta_\textrm{tsm}$, as shown in \figref{fig:rwtm_eff_over_dec_num_and_tsm_angle}.

    \item \textbf{Length Constraint:} The theoretical maximum length is not directly limited by efficiency, but by the total wire elongation, which complicates control. For simple binary operations like peristalsis, the length can be very long. However, for precise joint angle control, the allowable elongation, and thus the length, depends on the desired accuracy, wire material, and control system sophistication. 
    
    \item \textbf{Decoupled Joint Number Constraint:} To perform translation and rotation within a two-dimensional plane without bending the TSM, at least three decoupled joints are required. In more complex environments, the TSM may come into contact with its surroundings and bend, reducing efficiency; thus, additional joints become necessary. The required number of joints depends on the acceptable level of efficiency loss and the specific environmental conditions in which the system operates.

\end{itemize}

\subsection{\titlecap{Experimental Performance and Observations}}
Let us consider the locomotion experiment. The force generated by peristaltic motion was not necessarily large, but it was sufficient as an external force to deform the RWTM. This shows that the RWTM flexibly follows even weak external forces while transmitting a large tension of several hundred Newtons. Also, as the locomotion progresses, the angle of the decoupled joint changes, but not much bending is seen in the TSM. This proves that, as stated in \subsecref{subsec:RWTM_design}, the decoupled joints preferentially bore the bending angle over the TSM. As a result, the transmission efficiency remained high and stable (average 0.884, SD 0.016, as shown in \tabref{tab:eff_comp}), and there was no major change in the behavior of the distal mobile robot even when the configuration of the RWTM changed.

Let us consider the controller performance evaluation experiment. Overall, the simple controller, which did not include a friction/hysteresis model, was able to achieve a certain level of performance. Also, regarding torque discretization, only the NCC of the estimator improved. This is thought to be because, as seen in the preliminary test (\figref{fig:RWTM_single_exp}), the input and output tension patterns became more similar with discrete steps. However, this did not translate to better MSE, likely because the absolute tension values were not more accurate and the overall motion became rougher. The final end-effector position and orientation errors (0.42 m and 1.52 rad, respectively) highlight the impact of the unmodeled, complex friction within the transmission system.

In the detailed transmission experiment (\figref{fig:rwtm_eff_detail_exp}), a consistent efficiency offset was observed between the RWTM and the simple product of its parts, suggesting the friction at the interfaces between the TSM and decoupled joints.

Let us consider the underwater experiment. Since REWW-ARM has no electronic systems other than the motor-unit, it was expected that there would be no problem with its operation even when the distal mobile robot and the decoupled joint were completely submerged deep in the water. Also, even after being submerged for a while, no change in behavior or rust on the body was observed compared to before submersion. This is because the Vectran wire used \cite{vectran, vectran_about} is water-repellent, and most of the other metal parts were made of stainless steel or anodized aluminum. This suggests that the choice of materials for mechanical parts is wider than for electronic parts, and that they can be used in a variety of environments if selected appropriately. We also confirmed that locomotion and posture control other than those confirmed in the locomotion experiment, such as parallel translation and roll rotation, are also possible.

\section{\titlecap{Conclusion}}\label{sec:conclusion}
In this study, we proposed a new method for separating electronic devices from the operating environment, the Remote Wire Drive. As a proof-of-concept, we developed the Remote Wire-Driven robot REWW-ARM, and conducted its design, development, and experimental verification. The core technology, the "Remote Wire Transmission Mechanism (RWTM)," was shown to achieve highly efficient and stable transmission by combining decoupled joints and TSMs, maintaining an average efficiency of 0.884, a significant improvement over a pure TSM (0.669) over a wide range of motion. Furthermore, through the movements of the distal mobile robot's Gear-Coupled Dual-Axis Joint, Variable-Stiffness Contract Link, and Anchor-Gripper Integrated End-Effector,  we confirmed that autonomous locomotion and object manipulation of the robot through the RWTM are possible. In addition, through controller performance evaluation experiments and underwater experiments, we demonstrated the effectiveness and environmental resistance of the Remote Wire-Driven system. In the future, we expect more accurate state estimation and actuation with more advanced control architecture that deals with the complex friction phenomena, the application of the Remote Wire-Driven system to robots of various structures, and the application of those Remote Wire-Driven robots to extreme environments.
\bibliographystyle{IEEEtran}
\bibliography{bib}

\end{document}